\documentclass[final]{IEEEtran}

\usepackage{cite}
\usepackage[colorlinks, citecolor=blue]{hyperref}
\usepackage{times}
\usepackage{epsfig}
\usepackage{graphicx}
\usepackage{amsmath}
\usepackage{amssymb}
\usepackage{pifont}
\newcommand{\cmark}{\ding{51}}%
\newcommand{\xmark}{\ding{55}}%
\usepackage{subfigure}
\usepackage{dsfont}
\usepackage{amsthm}
\usepackage{mathrsfs}
\usepackage{algorithm}
\usepackage{algpseudocode}
\newtheorem{proposition}{Proposition}

\usepackage{xspace}

\makeatletter
\DeclareRobustCommand\onedot{\futurelet\@let@token\@onedot}
\def\@onedot{\ifx\@let@token.\else.\null\fi\xspace}

\def\eg{\emph{e.g}\onedot} 
\def\ie{\emph{i.e}\onedot}

\def\etal{\emph{et al}\onedot}
\makeatother

\begin{document}

\title{Conditional Generative ConvNets for \\Exemplar-based Texture Synthesis}

\author{Zi-Ming~Wang$^1$, Meng-Han Li$^2$,  Gui-Song~Xia$^{1}$\\
$^1${\em Wuhan University, Wuhan, China.}\\
$^2${\em Paris Descartes University, Paris, France.}\\
}

\maketitle

\begin{abstract}
	The goal of exemplar-based texture synthesis is to generate texture images that are visually similar to a given exemplar.
	Recently, 
	promising results have been reported by methods relying on convolutional neural networks (ConvNets) pretrained on large-scale image datasets. 
	However, 
	these methods have difficulties in synthesizing image textures with non-local structures and extending to dynamic or sound textures.	
	In this paper, 
	we present a {\em conditional generative ConvNet} (cgCNN) model which combines deep statistics and the probabilistic framework of generative ConvNet (gCNN) model. 
	Given a texture exemplar, 
	the cgCNN model defines a conditional distribution using deep statistics of a ConvNet, 
	and synthesize new textures by sampling from the conditional distribution.	
	In contrast to previous deep texture models, 
	the proposed cgCNN dose not rely on pre-trained ConvNets but learns the weights of ConvNets for each input exemplar instead.
	As a result, the cgCNN model can synthesize high quality dynamic, sound and image textures in a unified manner.
	We also explore the theoretical connections between our model and other texture models.
	Further investigations show that the cgCNN model can be easily generalized to texture expansion and inpainting.
	Extensive experiments demonstrate that our model can achieve better or at least comparable results than the state-of-the-art methods.
\end{abstract}

\section{Introduction}
Exemplar-based texture synthesis (EBTS) has been a dynamic yet challenging topic in computer vision and graphics for the past decades~\cite{heeger1995pyramid,Efros1999,Wei2000,Efros2001,zhu1998filters, portilla2000parametric,galerne2011random,Galerne_texton_noise_cgf2017,raad2017survey,xie2017synthesizing}, 
which targets to produce new samples that are visually similar to a given texture exemplar.
The main difficulty of EBTS is to efficiently synthesize texture samples that are not only perceptually similar to the exemplar,
but also able to balance the repeated and innovated elements in the texture.

To overcome this difficulty, two main categories of approaches have been proposed in the literature,
 \ie, patch-based methods~\cite{Efros1999,Wei2000,Efros2001,kaspar2015self} and methods relying on parametric statistical models~\cite{heeger1995pyramid,zhu1998filters,portilla2000parametric,galerne2011random,gatys2015texture}.
Given a texture exemplar, 
patch-based methods regard small patches in the exemplar as basic elements,
and generate new samples by copying pixels or patches from the exemplar to the synthesized texture under certain spatial constrains, 
such as Markovian property~\cite{Efros1999,kaspar2015self,Efros2001}. 
These methods can produce new textures with high visual fidelity to the given exemplar, 
but they often result in verbatim copies and few of them can be extended to dynamic texture, except~\cite{kwatra2003graphcut}.
Moreover, 
in contrast with their promising performance, 
they take less steps to understand the underlying process of textures. 
Whereas statistical parametric methods concentrate on exploring the underlying models of the texture exemplar, 
and new texture images can then be synthesized by sampling from the learned texture model. 
These methods are better at balancing the repetitions and innovations nature of textures, 
while they usually fail to reproduce textures with highly structured elements. 
It is worth mentioning that a few of these methods can be extended to sound textures~\cite{mcdermott2009sound} and dynamic ones~\cite{xia2014synthesizing}.
Some recent surveys on EBTS can be founded in~\cite{wei2009state,raad2017survey}. 

Recently, 
parametric models has been revived by the use of {\em deep neural networks}~\cite{gatys2015texture, ulyanov2017improved,sendik2017deep,liu2016texture}. 
These models employ deep ConvNets that are pretrained on large-scale image datasets instead of handcrafted filters as feature extractors,
and generate new samples by seeking images that maximize certain similarity between their deep features and those from the exemplar. 
Although these methods show great improvements over traditional parametric models,
there are still two unsolved or only partially solved problems:
1) It is difficult to extend these methods to other types of textures, 
such as dynamic and sound textures, since these methods rely on ConvNets pre-trained on large-scale datasets, such as ImageNet, which are difficult to obtain in video or sound domain.
2) These models can not synthesize textures with non-local structures,
as the optimization algorithm is likely to be trapped in local minimums where non-local structures are not preserved.
A common remedy is to use extra penalty terms, 
such as Fourier spectrum~\cite{liu2016texture} or correlation matrix~\cite{sendik2017deep},
but these terms bring in extra hyper-parameters and are slow to optimize.

In order to address these problems, 
we propose a new texture model named {\em conditional generative ConvNet} (cgCNN) by integrating deep texture statistics and the probabilistic framework of generative ConvNet (gCNN)~\cite{xie2016theory}.
Given a texture exemplar, 
cgCNN first defines an energy based conditional distribution using deep statistics of a trainable ConvNet, 
which is then trained by maximal likelihood estimation (MLE).
New textures can be synthesized by sampling from the learned conditional distribution.
Unlike previous texture models that rely on pretrained ConvNets,
cgCNN \emph{learns} the weights of the ConvNet for each input exemplar.
It therefore has two main advantages:
1) It allows to synthesize image, dynamic and sound textures in a unified manner. 
2) It can synthesize textures with non-local structures without using extra penalty terms,
as it is easier for the sampling algorithm to escape from local minimums.

We further present two forms of our cgCNN model, \ie 
the canonical cgCNN (c-cgCNN) and the forward cgCNN (f-cgCNN),
by exploiting two different sampling strategies.
We show that these two forms of cgCNN have strong theoretical connections with previous texture models.
Specifically,
c-cgCNN uses Langevin dynamics for sampling, 
and it can synthesize highly non-stationary textures.
While f-cgCNN uses a fully convolutional generator network as an approximated fast sampler, 
and it can synthesize arbitrarily large stationary textures.
We further show that \emph{Gatys' method~\cite{gatys2015texture} and TextureNet~\cite{ulyanov2017improved} are special cases of c-cgCNN and f-cgCNN respectively.}
In addition,
we derive a concise texture inpainting algorithm based on cgCNN,
which iteratively searches for a template in the uncorrupted region and synthesizes a texture patch according to the template.

Our main contributions are thus summarized as follows:
\begin{itemize}
	\item[-] We propose a new texture model named cgCNN which combines deep statistics and the probabilistic framework of gCNN model. 
	Instead of relying on pretrained ConvNets as previous deep texture models,
			the proposed cgCNN learns the weights of the ConvNet adaptively for each input exemplar.
			As a result, 
			cgCNN can synthesize high quality dynamic, sound and image textures in a unified manner.
	\item[-] We present two forms of cgCNN and show their effectiveness in texture synthesis and expansion: 			
	c-cgCNN can synthesize highly non-stationary textures without extra penalty terms, 
	while f-cgCNN can synthesize arbitrarily large stationary textures. 
	We also show their strong theoretical connections with previous texture models. 
	Note f-cgCNN is the first deep texture model that enables us to expand dynamic or sound textures.
	\item[-] We present a simple but effective algorithm for texture inpainting based on the proposed cgCNN. 
			To our knowledge, it is the first neural algorithm for inpainting sound textures.
	\item[-] Extensive experiments\footnote{All experiments can be found at \url{captain.whu.edu.cn/cgcnn-texture}.} in synthesis, expansion and inpainting of various types of textures using cgCNN. 
	We demonstrate that our model achieves better or at least comparable results than the state-of-the-art methods.
\end{itemize}

The rest of this paper is organized as follows:
Sec.~\ref{relatedwork} reviews some related works.
Sec.~\ref{Preliminaries} recalls four baseline models.
Sec.~\ref{Sec_our_model} details cgCNN's formulation and training algorithm, and provides some theoretical analysis to the models.
Sec.~\ref{Dynamic_and_Sound} uses cgCNN for the synthesis of various types of textures and adapts the synthesis algorithm to texture inpainting.
Sec.~\ref{experiments} presents results that demonstrate the effectiveness of cgCNN in synthesizing, expanding and inpainting all three types of textures.
Sec.~\ref{conclusion} draws some conclusions.

\section{Related Work}
\label{relatedwork}

One seminal work on parametric EBTS was made by Heeger and Bergen~\cite{heeger1995pyramid}, 
who proposed to synthesize textures by matching the marginal distributions of the synthesized and the exemplar texture. 
Subsequently, Portilla \etal~\cite{portilla2000parametric} extended this model by using more and higher-order measurements. 
Another remarkable work at that time was the FRAME model proposed by Zhu \etal~\cite{zhu1998filters}, 
which is a framework unifying the random field model and the maximum entropy principle for texture modeling. 
Other notable works include~\cite{galerne2011random,Galerne_texton_noise_cgf2017,Peyre2009}. 
These methods built solid theoretical background for texture synthesis,
but are limited in their ability to synthesize structured textures.

Recently, 
Gatys~\cite{gatys2015texture} made a breakthrough in texture modelling by using deep neural networks. 
This model can be seen as an extension of Portilla's model~\cite{portilla2000parametric}, 
where the linear filters was replaced by a pretrained deep ConvNet. 
Gatys' method was subsequently extended to style transfer~\cite{gatys2016image},
where the content image was force to have similar deep statistics with the style image.
In more recent works,
Gatys' method has been extended to synthesizing textures with non-local structures by using more constraints such as correlation matrix~\cite{sendik2017deep} and spectrum~\cite{liu2016texture}.
However,
such constraints bring in extra hyper-parameters that require manual tuning, 
and are slow to optimize~\cite{sendik2017deep} or cause spectrum like noise~\cite{liu2016texture}.
In contrast, 
our model can synthesize non-local structures without the aid of these constraints due to the effective sampling strategy.
In order to accelerate the synthesis process and synthesize larger textures than the input, 
Ulyanov \etal~\cite{ulyanov2017improved} and Johnson \etal~\cite{johnson2016perceptual} proposed to combine a fully convolutional generator with Gaty's model,
so that textures can be synthesized in a fast forward pass of the generator.
Similar to Ulyanov's model \etal~\cite{ulyanov2017improved},
our model also uses a generator for fast sampling and texture expansion.
In contrast to Gatys' method which relies on pretrained ConvNets,
Xie~\cite{xie2016theory} proposed a generative ConvNet (gCNN) model that can learn the ConvNet and synthesize textures simultaneously.
In subsequent works,
Xie~\cite{xie2018cooperative} proposed CoopNet by combining gCNN and a latent variable model.
This model was latter extended to video~\cite{xie2017synthesizing} and 3D shape~\cite{xie2018learning} synthesis.
Our model can be regarded as a combination of Gatys' method and gCNN,
as it utilizes the idea of deep statistics in Gatys' method and the probabilistic framework of gCNN.

Considering dynamic texture synthesis, 
it is common to use linear auto-regressive models~\cite{doretto2003dynamic,xia2014synthesizing} to model the appearance and dynamics. 
Later work~\cite{yang2018learning} compared these method quantitatively by studying the synthesizability of the input exemplars. 
Recent works leveraged deep learning techniques for synthesizing dynamic textures. 
For instance, 
Tesfaldet \etal~\cite{tesfaldet2018two} proposed to combine Gatys' method~\cite{gatys2015texture} with an optic flow network in order to capture the temporal statistics. 
In contrast, 
our model does not require the aid of other nets, 
as our model is flexible to use spatial-temporal ConvNets for spatial and temporal modelling.

As for sound texture synthesis,
classic models~\cite{mcdermott2009sound} are generally based on wavelet framework and use handcrafted filters to extract temporal statistics.
Recently,
Antognini \etal~\cite{antognini2019audio} extended Gatys' method to sound texture synthesis by applying a random network to the spectrograms of sound textures.
In contrast,
our model learns the network adaptively instead of fixing it to random weights,
and our model is applied to raw waveforms directly.

The texture inpainting problem is a special case of image or video inpainting problem, 
where the inpainted image or video is assumed to be a texture.
Igehy~\cite{igehy1997image} transferred Heeger and Bergen's texture synthesis algorithm~\cite{heeger1995pyramid} to an inpainting algorithm.
Our inpainting algorithm shares important ideas with Igehy's method~\cite{igehy1997image},
as we also adopt an inpainting by synthesizing scheme.
Other important texture inpainting methods include conditional Gaussian simulation~\cite{galerne2017texture} and PatchMatch based methods~\cite{liu2013exemplar-based, barnes2009patchmatch}.

\section{Preliminaries}
\label{Preliminaries}
This section recalls several baseline models on which cgCNN is built.
The theoretical connections between these model and cgCNN will be discussed in Sec.~\ref{Sec_our_model}.

Given a RGB-color image texture exemplar $f_0 \in \mathbb{R}^{H \times W \times 3}$,
where $H$ and $W$ are the height and width of the image,
texture synthesis targets to generate new samples $f\in \mathbb{R}^{H \times W \times 3}$ that are visually similar to $f_0$.

\paragraph{\bf Gatys' method~\cite{gatys2015texture}}

\emph{Gatys' method} uses a pretrained deep ConvNet as a feature extractor. 
For an input texture exemplar, 
the Gram matrices of feature maps at selected layers are first calculated. 
New texture samples are then synthesized by matching the Gram matrices of the synthesized textures and the exemplar.

Formally,
Gatys' method tries to solve the following optimization problem:
\begin{equation}
	\label{Gatys_model}
	\min_{f} L_{G}(f,\, f_0).
\end{equation}
The objective function $L_{G}$ is defined as:
\begin{equation}
\label{Gatys}
L_{G}(f,\, f_0)=\sum_{l} \big\|\mathbf{G}\big(\mathcal{F}^{(l)}(f) \big) - \mathbf{G}\big(\mathcal{F}^{(l)}(f_0) \big)\big\|_F,
\end{equation} 
where $\mathcal{F}$ is a pretrained ConvNet, 
$\mathcal{F}^{(l)}$  is the feature map at layer $l$, 
and $\|\cdot\|_F$ is the Frobenius norm.  
$\mathbf{G}$ is the Gram matrix defined as:
\begin{equation}\label{Gram}
\mathbf{G} = F^{T} F \, \in \mathbb{R}^{C \times C}, 
\end{equation}
where $F = \mathcal{F}^{(l)}(f) \in \mathbb{R}^{N \times C}$ is a feature map with $C$ channels and $N$ elements in each channel.

This model is trained by gradient descent using back propagation.
Each step follows
\begin{equation}\label{gatys_Langevin}
f_{t+1} = f_{t}- \epsilon  \frac{\partial L_{G}(f_{t},\, f_0)}{\partial f_{t}},
\end{equation}
where $\epsilon$ is the learning rate.

\paragraph{\bf TextureNet~\cite{ulyanov2017improved}}

\emph{TextureNet} is a forward version of Gatys' method. 
It learns a generator network $g_{\mathbf \theta}$ with trainable weights $\mathbf \theta$, 
which maps a sample of random noise $z \sim \mathcal{N}(0, I)$ to a local minimum of Eqn.~\eqref{Gatys}.
This amounts to solve the following optimization problem:

\begin{equation}
	\label{texture_net}
	\min_{\theta} \mathbb{E}_{z \sim \mathcal{N}(0, I)} \Big(L_{G}(g_{\mathbf \theta}(z),\, f_0)\Big).
\end{equation}
$g_{\mathbf \theta}$ is trained by gradient decent with approximate gradients:

\begin{equation}
	\label{texture_net_train}
	\frac{\partial L_{TN}(\theta)}{\partial \theta} = \frac{1}{N} \sum_{i}   \frac{\partial L_{G}(g_{\mathbf \theta}(z_i),\, f_0)}{\partial \theta},
\end{equation}
where $z_1,...,z_N$ are $N$ samples from $\mathcal{N}(0, I)$.

\paragraph{\bf Generative ConvNet (gCNN)~\cite{xie2016theory}}

\emph{gCNN} is defined on a more general setting.
It aims to estimate the underlying distribution of a set of images $\{f_k\}_{k=0}^K$ 
and generate new images by sampling from this distribution.
In our work, 
we only consider the specific case where the input set contains only one image $f_0$, \ie $K=0$,
and $f_0$ is a stationary texture exemplar. 

gCNN defines a distribution of $f$ in image space:
\begin{equation}\label{gCNN_model}
\mathbf{P}(f; \,\alpha) = \frac{1}{Z(\alpha)} e^{-E_{g}(f; \, \alpha)},
\end{equation}
where $Z(\alpha)=\sum_f e^{-E_{g}(f; \, \alpha)}$ is the normalization factor. 
$E_{g}(f; \,\alpha)$ is the energy function defined by
\begin{equation}
	\label{gCNN_energy}
	E_{g}(f; \,\alpha) = - \sum \mathcal{F}_{\alpha}(f),
\end{equation}
where $\mathcal{F}_{\alpha}$ is the output of a ConvNet with learnable weights $\alpha$. 
gCNN is trained by maximum likelihood estimation (MLE).

\paragraph{\bf CoopNet~\cite{xie2018cooperative}} 

\emph{CoopNet} extends gCNN by combining gCNN with a latent variable model~\cite{han2017alternating} which takes the form of 
\begin{equation}
	f = g_{l}(z; \theta)+ \sigma ,   z \sim \mathcal{N}(0, I), \sigma \sim \mathcal{N}(0, \delta ^2),
\end{equation}
where $g_{l}$ is a forward ConvNet parametrized by $\theta$ and $f$ is the synthesized image.

The $g_{l}$ is trained by MLE,
which is to iterate the following four steps:
\begin{itemize}
	\item[1)] Generate samples $f = g_{l}(z; \theta)$ using random $z\sim\mathcal{N}(0, I)$.
	\item[2)] Feed $f$ to gCNN, run $l_{d}$ steps of Langevin dynamics for $f$: $\hat{f}=f -\epsilon \frac{\partial}{\partial f} E_{g}(f) + noise$.
	\item[3)] Run $l_{g}$ steps of Langevin dynamics for $z$: $\hat{z}=z - \epsilon \frac{\partial}{\partial z} ||\hat{f} - g_{l}(z; \theta)||^2 + noise$.
	\item[4)] Update $\theta$ using gradient descent: $\theta$ = $\theta - \epsilon \frac{\partial}{\partial \theta} ||\hat{f} - g_{l}(\hat{z}; \theta)||^2$
\end{itemize}

\section{Conditional generative ConvNet}
\label{Sec_our_model}
In this section, 
we first present the definition of our conditional generative ConvNet (cgCNN) model, 
and then explore two forms of cgCNN, \ie the canonical cgCNN (c-cgCNN) and the forward cgCNN (f-cgCNN).
Finally,
we conclude this section with some theoretical explanations of cgCNN.

\subsection{Model formulation}
Let $f_0$ represent an image, dynamic or sound texture exemplar, and note that the shape of $f_0$ depends on its type.
Specifically,
$f_0 \in \mathbb{R}^{H \times W \times T \times 3}$ represents a dynamic texture exemplar; 
$f_0 \in \mathbb{R}^{H \times W \times 3}$ represents an image texture exemplar,
and $f_0 \in \mathbb{R}^{ T }$ represents a sound texture exemplar,
where $H \times W$ and $T$ are spatial and temporal sizes.

Given a texture exemplar $f_0$, 
cgCNN defines a conditional distribution of synthesized texture $f$:
\begin{equation}\label{our_model}
	\mathbf{P}(f \,|\, f_0; \,w) = \frac{1}{Z(w)} e^{-E_{cg}(f,\, f_0; \,w)},
\end{equation}
where $Z(w)=\sum_f e^{-E_{cg}(f,f_0; w)} $ is the normalization factor. 
$E_{cg}(f,\, f_0; \,w)$ is the energy function which is supposed to capture the visual difference between $f$ and $f_0$ by assigning lower values to $f$'s that are visually closer to $f_0$. 
As an analogue to $L_{G}$,
we define $E_{cg}$ by 
\begin{equation}
	\label{our_energy}
		E_{cg}(f,\, f_0; \,w) = \sum_l \| \mathbf{S}(\mathcal{D}^{(l)}_{w}(f_0)) - \mathbf{S}(\mathcal{D}^{(l)}_{w}(f)) \|_F,	
\end{equation}
where $\mathcal{D}_{w}$ is a deep network with learnable weight $w$, 
and  $\mathcal{D}_{w}^{(l)}$ is the feature maps at the $l$-th layer. 
$\mathbf S$ is a statistic measurement, 
such as \eg Gram matrix $\mathbf{G}$ defined in Eqn.~\eqref{Gram}.
We also test spatial mean vector $\mathbf{M}$ as an alternative measurement in our experiment section. 
For simplicity,
in the rest of this paper,
we denote $\mathbf{P}(f \,|\, f_0; \,w)$ by $\mathbf{P}(w)$ when the meaning is clear from the text.

\subsection{Training and sampling}

The objective of training cgCNN is to estimate the conditional distribution $\mathbf{P}(w)$ using only one input data $f_0$.
This is achieve by minimizing the KL divergence between the empirical data distribution, 
which is a Kronecker delta function $\delta_{f_0}$,
and the estimated distribution $\mathbf{P}(w)$.
The KL divergence $KL(w)$ can be written as:
\begin{align}
	KL(w) &= KL(\delta_{f_0} || \mathbf{P}(w)) \nonumber \\
	&= -H(\delta_{f_0})  - \log \mathbf P(f_0\,|\,f_0; \,w) \nonumber \\
													&= -\log \mathbf P(f_0\,|\,f_0; \,w)  \nonumber \\
													&= \log Z(w),
\end{align}
where $H(\cdot)$ denotes the entropy
and $H(\delta_{f_0})=0$.

Note that minimizing $KL(w)$ is equivalent to MLE,
where the log-likelihood $L(w)$ is defined as the log-likelihood of the input $f_0$ given $f_0$ itself as the condition:
\begin{equation}
L(w) = \log \mathbf P(f_0\,|\,f_0; \,w) = -\log Z(w).
\end{equation}
For the consistency of notation,
in the rest of this paper,
we use $KL(w)$ instead of $L(w)$ as the objective function.

The gradient of $KL(w)$ can be written as follows:
\begin{equation}\label{our_gradient_L}
 \frac{\partial KL(w)}{\partial w} = \mathbb{E}_{f \sim \mathbf{P}(w)}\Big(-\frac{\partial E_{cg}(f, \, f_0; \, w)}{\partial w} \Big).
\end{equation}
Note that the expectation term $\mathbb{E}_{f \sim \mathbf{P}(w)}(\cdot)$ in Eqn.~\eqref{our_gradient_L} is analytical intractable,
and has to be approximated by the Monte Carlo method. 
Suppose we have $K$ samples $f^{(1)},...,f^{(K)}$ drawn from $\mathbf{P}(w)$,
the gradient of $KL(w)$ can be approximated as:
\begin{equation}\label{our_gradient_L2}
	\frac{\partial KL(w)}{\partial w} = -\frac{1}{K}\sum_{k=1}^{K} \frac{\partial E_{cg}(f^{(k)}, \, f_0; \,w)}{\partial w}.
\end{equation}
We can then minimize $KL(w)$ using gradient decent according to Eqn.~\eqref{our_gradient_L2}.

Therefore, 
the key of training cgCNN is sampling from $\mathbf{P}(w)$.
We use 1) Langevin dynamics and 2) a generator net for sampling, 
which lead to c-cgCNN and f-cgCNN respectively.

\subsubsection{c-cgCNN}

c-cgCNN uses Langevin dynamics to sample from $\mathbf{P}(w)$.
Specifically, 
starting from a random noise $f$, 
it uses the following rule to update $f$:
\begin{equation}\label{our_Langevin}
f_{t+1} = f_{t}- \frac{\epsilon^2}{2}  \frac{\partial E_{cg}(f_t, \, f_0; \, w)}{\partial f_t} + \epsilon N_t,
\end{equation}
where $f_{t}$ is a sample at step $t$, 
$\epsilon$ is the step size, 
and $N_t \sim \mathcal{N}(0, 1)$ is a Gaussian noise. 
A training algorithm for c-cgCNN can be derived by combining Langevin sampling in Eqn.~\eqref{our_Langevin} and approximated gradient in Eqn.~\eqref{our_gradient_L2}.
Starting from a random noise $f$,
the algorithm iteratively goes through $\mathcal{D}$-learning step and Langevin sampling step: 
\begin{itemize}
	\item[-]  {\bf Langevin sampling:} draw samples using Langevin dynamics according to Eqn.~\eqref{our_Langevin}.
	\item[-] {\bf $\mathcal{D}$-learning:} update network $\mathcal{D}$ using approximated gradient according to Eqn.~\eqref{our_gradient_L2}.
\end{itemize}
The detailed training process is presented in Alg.~\ref{our_algorithm_1}.

\begin{algorithm}[h]
	\caption{Training and sampling from c-cgCNN}\label{our_algorithm_1}
	\begin{algorithmic}
	\Require a texture exemplar $f_0$, Langevin sampling steps $N$, training steps $T$, and the number of synthesis textures $K$.
	\Ensure Synthesized textures $\hat{f}$ and learned network $\mathcal{D}_w$.
	\vspace{5mm}
	\State {\bf Initialize} $t \leftarrow 0$; \; $\hat{f}^{(k)} \leftarrow \mathcal{N}(0, \, 1), \, k=1, \ldots, K$; \; 
	\For{$t= 1, \ldots, T$}
		  \State {\bf Langevin sampling:} Run $N$ Langevin steps for all $K$ textures $\{\hat{f}^{(k)}\}_{k=1}^K$. Each step follows Eqn.~\eqref{our_Langevin}. 
       	 \State {\bf $\mathcal{D}$-learning:} Update network $\mathcal{D}_w$: $w \leftarrow w - \frac{\partial KL(w)}{\partial w}$, where $\frac{\partial KL(w)}{\partial w}$ is calculated according to Eqn.~\eqref{our_gradient_L2}.
        \EndFor
	\end{algorithmic}
\end{algorithm}

\subsubsection{f-cgCNN}
The Langevin dynamics used in c-cgCNN is slow,
and may be the bottleneck of the Alg.~\ref{our_algorithm_1}.
As an alternative, 
we may also use a generator net as a fast approximated sampler of $\mathbf{P}(w)$. 
Specifically,
we introduce a generator network $\mathcal{G}_{\theta}$ with learnable weights $\theta$, 
which maps the normal distribution $\mathcal{N}(0, I)$ to a parametrized distribution $\mathbf{Q}(\theta)$.
The training object is to match $\mathbf{Q}(\theta)$ and $\mathbf{P}(w)$, 
so that samples of $\mathbf{P}(w)$ can be approximated by samples of $\mathbf{Q}(\theta)$.
In other words,
when $\mathcal{G}_{\theta}$ is trained,
approximated samples of $\mathbf{P}(w)$ can be drawn by forwarding a noise $z\sim\mathcal{N}(0, I)$ through network $\mathcal{G}_{\theta}$,
which is much faster than Langevin dynamics in Eqn.~\eqref{our_Langevin}.
Formally,
network $\mathcal{G}_{\theta}$ is trained by minimizing the KL divergence between $\mathbf{Q}(\theta)$ and $\mathbf{P}(w)$:
\begin{align}
	\label{train_G}
	KL(\theta)&=KL(\mathbf{Q(\theta)}||\mathbf{P}(w)) \nonumber \\
				&=-H(\mathbf{Q}(\theta))-\mathbb{E}_{f \sim \mathbf{Q}(\theta)}log\mathbf{P}(f \,|\, f_0; \,w) \nonumber \\
				&=-H(\mathbf{Q}(\theta))+\mathbb{E}_{f \sim \mathbf{Q}(\theta)}E_{cg}(f, \, f_0; \, w) + const.
\end{align}

The first term $H(\mathbf{Q}(\theta))$ in Eqn.~\eqref{train_G} is the entropy of distribution $\mathbf{Q}(\theta)$,
which is analytical intractable.
Following TextureNet~\cite{ulyanov2017improved},
we use Kozachenko-Leonenko estimator~\cite{kozachenko1987sample} (KLE) to approximate this term.
Given $K$ samples $f^{(1)},...,f^{(K)}$ drawn from $\mathbf{Q}(\theta)$, 
KLE is defined as:
\begin{equation}
	KLE(\theta) = \sum_{0<i,j<K}||f^{(i)} - f^{(j)}||_F.
\end{equation}

The second term in Eqn.~\eqref{train_G} is an expectation of our energy function $E_{cg}$. 
It can be approximated by taking average over a batch of samples of $\mathbf{Q}(\theta)$.

Now, 
since both terms in Eqn.~\eqref{train_G} can be approximated,
the gradient of $KL(\theta)$ can be calculated as:
\begin{equation} \label{g_gradient}
	\frac{\partial KL(\theta)}{\partial \theta} = - \frac{\partial KLE(\theta)}{\partial \theta} + \frac{1}{K}\sum_{k=1}^{K} \frac{\partial E_{cg}(\mathcal{G}_{\theta}(z_k), \, f_0; \,w)}{\partial \theta} , 
\end{equation}
where $z_1, z_2, ... ,z_K$ are $K$ samples drawn from $\mathcal{N}(0, I)$.

The complete training algorithm of f-cgCNN can be derived by training network $\mathcal{G}$ and $\mathcal{D}$ jointly.
Formally,
the goal is to match three distributions: $\mathbf{Q}(\theta)$, $\mathbf{P}(w)$ and $\delta_{f_0}$ by optimizing the following objective function,
\begin{equation}
	\min_{\mathcal{D}_{w}}\min_{\mathcal{G}_{\theta}} KL(\theta) + KL(w).
\end{equation}
To achieve this goal,
f-cgCNN is trained by iteratively going through the following three steps: 
\begin{itemize}
	\item[-] {\bf $\mathcal{G}$-synthesis:} generate $K$ samples using network $\mathcal{G}$.
	\item[-] {\bf $\mathcal{D}$-learning:} update network $\mathcal{D}$ using approximated gradient according to Eqn.~\eqref{our_gradient_L2}.
	\item[-]  {\bf $\mathcal{G}$-learning:} update network $\mathcal{G}$ using approximated gradient according to Eqn.~\eqref{g_gradient}.
\end{itemize}

The detailed algorithm is presented in Alg.~\ref{our_algorithm_2}.

\begin{algorithm}[h]
	\caption{Training f-cgCNN}\label{our_algorithm_2}
	\begin{algorithmic}
	\Require a texture exemplar $f_0$, training steps $T$, batch size $K$.
	\Ensure learned network $\mathcal{D}_w$ and learned network $\mathcal{G}_\theta$.
	\vspace{5mm}
	\State {\bf Initialize} $t \leftarrow 0$; \;  
	\For{$t= 1, \ldots, T$}
		\State {\bf $\mathcal{G}$-synthesis:} Sample a batch of $K$ noise $z_1$, ... $z_K$ from $\mathcal{N}(0, I)$, then generate $\mathcal{G}_{\theta}(z_1)$, ... , $\mathcal{G}_{\theta}(z_K)$. 
       	\State {\bf $\mathcal{D}$-learning:} Update network $\mathcal{D}_w$: $w \leftarrow w - \frac{\partial KL(w)}{\partial w}$, where $\frac{\partial KL(w)}{\partial w}$ is calculated according to Eqn.~\eqref{our_gradient_L2}.

		\State {\bf $\mathcal{G}$-learning:} Update network $\mathcal{G}_\theta$: $\theta \leftarrow \theta - \frac{\partial KL(\theta)}{\partial \theta}$, where $\frac{\partial KL(\theta)}{\partial \theta}$ is calculated according to Eqn.~\eqref{g_gradient}.

        \EndFor
	\end{algorithmic}
\end{algorithm}

\subsection{Theoretical understanding of cgCNN}
We present some theoretical understandings of cgCNN by relating it to other neural models.
We first point out cgCNN is conceptually related to GAN~\cite{goodfellow2014generative} as it can be written in a min-max adversarial form.
Then we show that:
1) c-gCNN and f-cgCNN are generalizations of Gatys' method~\cite{gatys2015texture} and TextureNet~\cite{ulyanov2017improved} respectively. 
2) c-cgCNN is a variation of gCNN~\cite{xie2016theory} with extra deep statistics, and the forward structures in f-cgCNN and CoopNet are consistent.
The main properties of these models are summarized in Tab.~\ref{model_comparison}.

\subsubsection{An adversarial interpretation}
The adversarial form of f-cgCNN can be written as:
\begin{equation}
	\label{Our2}
	\min_{\mathcal{G}_\theta} \max_{\mathcal{D}_w}  \mathbb{E}_{z\sim \mathcal{N}(0, I)} [ E_{cg}(\mathcal{G}_{\theta}(z) , \, f_0; \,w) ].
\end{equation}
This adversarial form has an intuitive explanation: 
network $\mathcal{G}_\theta$ tries to synthesize textures that are more visually similar to the input exemplar, 
and network $\mathcal{D}_w$ tries to detect the differences between them. 
The training process ends when the adversarial game reaches an equilibrium.
Similarly,
we have the min-max form of c-cgCNN:
\begin{equation}
	\label{Our1}
	\min_{f} \max_{\mathcal{D}_w} E_{cg}(f, \, f_0; \,w),
\end{equation}
where the synthesized texture $f$ plays the role that is played by $\mathcal{G}_\theta$ in f-cgCNN.

\subsubsection{Related to Gatys' method and TextureNet}
\label{model_compare}
It is easy to see that c-cgCNN is a generalization of Gatys' method with an extra step to learn the network $\mathcal{D}$.
Because if we fix network $\mathcal{D}$ to be a pretrained ConvNet with weights $w_0$ in Eqn.~\eqref{Our1},
c-cgCNN becomes $\min_{f}  E_{cg}(f, \, f_0; \,w_0)$,
which is exactly Gatys' method defined in Eqn.~\eqref{Gatys_model}.
Furthermore,
since f-cgCNN and TextureNet are built on c-cgCNN and Gatys' method respectively,
and they use the same forward structures,
we can conclude that f-cgCNN is a generalization of TextureNet as defined in Eqn.~\eqref{texture_net}.
In summary,
we have the following proposition:
\begin{proposition}
Gatys' method and TextureNet are special cases of c-cgCNN and f-cgCNN respectively, 
where the net $\mathcal{D}$ is fixed to be a pretrained ConvNet. 
\end{proposition}
Comparing to Gatys' method, 
samples of c-cgCNN are less likely to be trapped in local minimums for too long, 
because the $\mathcal{D}$-learning step always seeks to increase the energy of current samples.
For example,
if $f_{t}$ is a local minimal at step $t$,
the subsequent $\mathcal{D}$-learning step will increase $f_{t}$'s energy,
thus the energy of $f_{t}$ may be higher than its neighborhood at the beginning of step $t+1$,
and the Langevin steps will sample $f_{t+1}$ different from $f_{t}$.
In our experiments,
we find this property enables us to synthesize highly structured textures without extra penalty terms.

Unlike TextureNet and Gatys' method,
both c-cgCNN and f-cgCNN can synthesize other types of textures besides image texture,
because they do not rely on pretrained ConvNets.
In addition,
thanks to the their forward structures,
both f-cgCNN and TextureNet can synthesize textures that are larger than the input.

\subsubsection{Related to gCNN and CoopNet}
In general,
c-cgCNN can be regarded as a variation of gCNN in texture synthesis.
It should be noticed that the energy $E_g$ defined in gCNN dose not involve any deep statistics,
hence it can be used to synthesis both texture and non-texture images, such as human faces.
However,
the energy $E_{cg}$ defined in cgCNN incorporates deep statistics (Gram matrix or mean vector) specifically designed for texture modelling,
hence it is more powerful in texture synthesis but can not handle non-texture images.

CoopNet uses a latent variable model as the forward structure to accelerate the Langevin dynamics in gCNN. 
Note that the forward structures in CoopNet and f-cgCNN are consistent,
as they both seek to learn the distribution defined by their respective backward structures,\ie gCNN and cgCNN.
Furthermore,
they are equivalent in a special setting as stated in the following proposition.
\begin{proposition}
If we 1) disable all noise term in Langevin dynamics,
2) set $l_{d}=1, l_{g}=0$ in CoopNet,
and 3) discard the entropy term in f-cgCNN, 
the forward structures in CoopNet and f-cgCNN become equivalent.
\end{proposition}
In this setting,
denote the output of the latent variable model in gCNN as $f$,
then the target $\hat{f}$ is defined as $l_{d}=1$ step Langevin dynamics starting from $f$, \ie
$\hat{f}=f- \frac{\partial E_{g} }{\partial f}$.
Training $g_l$ amounts to minimize the objective function $||\hat{f}-f||^2 $ via gradient descent. 
Note the gradient of the objective function can be calculated as $\frac{\partial E_{g} }{\partial f} \frac{\partial f}{\partial \theta}$,
which is exactly back-propagation for minimizing $E_{g}$ according to the chain rule.
Because the generator net in f-cgCNN is also trained using back-propagation,
it is clear that the forward structures in CoopNet and f-cgCNN are equivalent.

All of cgCNN, CoopNet and gCNN can synthesize various types of textures.
However,
unlike f-cgCNN whose synthesis step is a simple forward pass,
the synthesis step of CoopNet involves several Langevin steps of gCNN, 
it is therefore difficult to expand textures using CoopNet.
  
\begin{table*}[ht!]
	\scriptsize
	\begin{center}
	  \caption{Comparison among several related models. 
	  Comparing to Gatys' method and TextureNet, 
	  cgCNN can synthesize more types of textures besides image textures.
	  Comparing to gCNN and CoopNet,
	  cgCNN incorporates extra multi-scale deep statistics which are more suitable for texture modelling.}
	  \label{model_comparison}
	  \begin{tabular}{r c c  p{1.4cm} p{1.2cm} p{1.2cm} p{1.1cm} p{1.2cm}} 
		  \hline
		  \textbf{Model} & \textbf{Forward structure} & \textbf{Backward structure} & \textbf{Multi-scale statistics} & \textbf{Dynamic texture synthesis}& \textbf{Sound texture synthesis} & \textbf{Texture expansion} & \textbf{Fast sampling}\\
		\hline
	  \centering
		Gatys'~\cite{gatys2015texture} & \line(1,0){4} &  pretrained ConvNet &\cmark & \xmark & \xmark & \xmark & \xmark \\
		TextureNet~\cite{ulyanov2017improved} & generator & pretrained ConvNet &\cmark & \xmark & \xmark & \cmark& \cmark \\
		gCNN~\cite{xie2016theory} & \line(1,0){4} & gCNN &  \xmark &\cmark & \line(1,0){4} &  \xmark& \xmark\\
		CoopNet~\cite{xie2018cooperative} & latent variable model & gCNN &\xmark & \cmark & \line(1,0){4} & \xmark & \cmark\\
		\hline
		c-cgCNN \textbf{(Ours)} & \line(1,0){4} & cgCNN & \cmark &\cmark & \cmark & \xmark& \xmark \\
		f-cgCNN \textbf{(Ours)} & generator & cgCNN & \cmark &\cmark & \cmark & \cmark & \cmark \\
		\hline
	  \end{tabular}
	\end{center}
	\vspace{-4mm}
\end{table*}

\section{Synthesize and inpaint image, dynamic and sound textures}

\label{Dynamic_and_Sound}
\subsection{Texture synthesis}
In our model, 
we use the same training algorithms described in Alg.~\ref{our_algorithm_1} and Alg.~\ref{our_algorithm_2} and statistics (Gram matrix and mean vector) for all types of textures.
Therefore,
in order to synthesize different types of textures,
we only need to modify the network dimensions accordingly,
and all other settings remain the same.

\subsubsection{Image texture synthesis}
Similar to previous image texture models~\cite{gatys2015texture,liu2016texture},
we use a 2-dimensional ConvNet to capture the spatial statistics.
Multi-scale statistics are captured in different layers in the networks.

\subsubsection{Dynamic texture synthesis}
Dynamic textures can be regarded as image textures with an extra temporal dimension.
Therefore, 
we simply use 3-dimensional spatial-temporal convolutional layers in cgCNN to capture the spatial appearances and the temporal dynamics simultaneously.
In other words,
unlike the methods~\cite{tesfaldet2018two, doretto2003dynamic} that model spatial and temporal statistics independently,
our model treats them equally by regarding a clip of dynamic texture as a spatial-temporal volume,
in which both the spatial and the temporal dimensions are stationary.

\subsubsection{Sound texture synthesis}
Sound textures can be regarded as a special case of dynamic textures, 
where spatial dimensions are not considered.
However,
modelling sound texture is not a simple task, 
because the sampling frequency of sound textures ($\sim$10 kHz) is usually far higher than that of dynamic textures ($\sim$10 Hz).
As a result, 
sound textures show more complicated long-range temporal dependencies and multi-scale structures than dynamic textures.

In our model,
we simply use 1-dimensional temporal convolutional layers in cgCNN to extract temporal statistics.
We use atrous~\cite{chen2017deeplab} convolutions to ensure large receptive fields, 
which enable us to learn long-range dependencies.
Unlike Antognini's model~\cite{antognini2019audio} which applies fixed random ConvNets to the spectrograms,
our model learns the ConvNet using raw waveforms.

\subsection{Texture inpainting}
\label{Section_inpainting}
As a proof of concept, 
we present a simple algorithm for texture inpainting based on our texture synthesis algorithm described in Alg.~\ref{our_algorithm_1}.

Given an input texture $f_0$ with a corrupted region $\Omega$,
the texture inpainting problem is to fill $\Omega$ so that the inpainted texture appears as natural as possible.
In other words,
$\overline{\Omega}$ must be visually close to at least one patch in the uncorrupted region $f_0 \setminus \Omega$,
where $\overline{\Omega}$ is the corrupted region with its border.

Our texture synthesis algorithm described in Alg.~\ref{our_algorithm_1} can be easily generalized to a texture inpainting algorithm, 
which iteratively searches for a template in $f_0 \setminus \Omega$ and updates $\Omega$ according to the template.
Specifically,
our method iterates a searching step and a synthesis step.
In the searching step,
we first measure the energy $E_{cg}(\phi, \overline{\Omega};w)$ between $\overline{\Omega}$ and all candidate patches $\phi \in f_0 \setminus \Omega$.
Then we select the patch $\phi^*$ with the lowest energy to be the template.
In the synthesis step,
we update $\Omega$ according to template $\phi^*$ using Alg.~\ref{our_algorithm_1}.
It is obvious that this algorithm ensures the inpainted region is visually similar to at least one patch (\eg the template) in the uncorrupted region.

In the searching step,
we use grid search to find the template $\phi^*$.
Note the template $\phi^*$ can also be assigned by the user~\cite{igehy1997image}.
It is possible to replace the grid search by more advanced searching techniques such as PatchMatch~\cite{barnes2009patchmatch},
and use gradient penalty~\cite{darabi2012image} or partial mask~\cite{igehy1997image} to ensure a smooth transition near the border of $\Omega$.
However,
these contradict the purpose of this algorithm,
which is to show the effectiveness of the proposed c-cgCNN method by combining it with other simplest possible methods.

The detailed inpainting algorithm is presented in Alg.~\ref{our_algorithm_1}.
\begin{algorithm}[h]
	\caption{Texture inpainting using c-cgCNN}\label{our_algorithm_3}
	\begin{algorithmic}
	\Require a texture exemplar $f$ with corrupted region $\Omega$. Langevin sampling step $N$, searching step $T$, updating step~$S$, network $\mathcal{D}_w$.
	\Ensure inpainted image $\tilde{f}$ and learned network $\mathcal{D}_w$.
	\vspace{5mm}
	\State {\bf Initialize} $t \leftarrow 0$; \; $\Omega \leftarrow 0$
	\For{$t= 1, \ldots, T$}
	\State ({\bf Template searching})
	\State  Find the patch $\phi^*$ with the lowest energy $E_{cg}(\phi, \overline{\Omega};w)$ amongst all patches $\phi \in f_0 \setminus \Omega$. Set $\phi^*$ to be the template.
	\State ({\bf c-cgCNN synthesis with exemplar $\phi^*$})
	\For{$s= 1, \ldots, S$}
			\State  Run $N$ Langevin steps for $\Omega$. Each step follows Eqn.~\eqref{our_Langevin}. 
			\State  Update network $\mathcal{D}_w$ $w$: $w \leftarrow w - \frac{\partial KL(w)}{\partial w}$, where $\frac{\partial KL(w)}{\partial w}$ is calculated according to Eqn.~\eqref{our_gradient_L2}.
		\EndFor
	\EndFor
	\end{algorithmic}
\end{algorithm}

\section{Experiments and Analysis}
\label{experiments}
In this section, 
we evaluate the proposed cgCNN model and compare it with other texture models.
We first perform self evaluations of c-cgCNN in Sec.~\ref{Bounded_experiment}-Sec.~\ref{Ablation_experiment}.
Specifically,
we investigate several key aspects of c-cgCNN including the influence of bounded constraints and the diversity of synthesis results.
We also carry out two ablation studies concerning the network structure and the training algorithm respectively.
Then we evaluate the performance of c-cgCNN and f-cgCNN in texture synthesis and expansion by comparing them with other theoretically related or the state-of-the-art methods in Sec.~\ref{Synthesis_experiment}-Sec.~\ref{Expansion_experiment}.
We finally evaluate our texture inpainting method in Sec.~\ref{Inpainting_experiment}.

\subsection{Experimental setting}
\paragraph{Exemplar} The image exemplars are collected from DTD dataset~\cite{cimpoi14describing} and Internet, 
and all examples are resized to $(256,256)$. 
The dynamic texture exemplars are adopted from~\cite{tesfaldet2018two},
where each video has 12 frames ane each frame is resized to $(128, 128)$.
We use sound textures that were used in~\cite{mcdermott2009sound},
which are recorded at $22050Hz$.
For our experiments, 
we clip the first $50000$ sample points (about 2 seconds) of each audio as exemplars.

\begin{figure}[htb!]
	\vspace{-3mm}
	\centering
	\includegraphics[width=.8\linewidth]{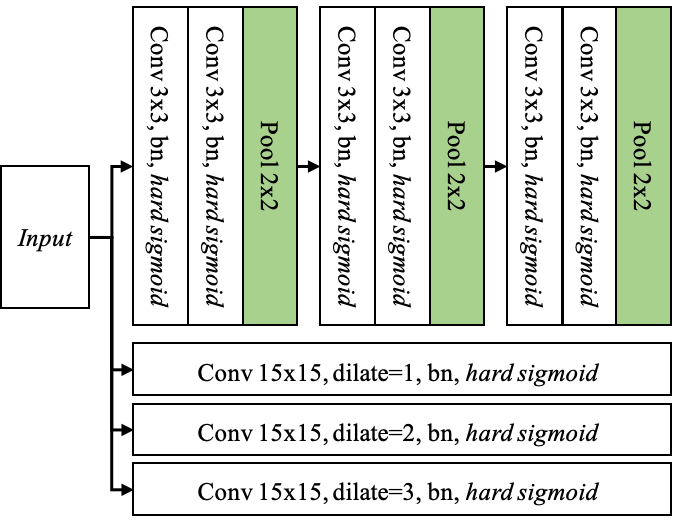}
	\caption{The network $\mathcal{D}$ used in cgCNN.}
	\label{network}
	\vspace{-2mm}
\end{figure}

\paragraph{Network architecture} 
The network $\mathcal{D}$ used in cgCNN is shown in Fig.~\ref{network}. 
It consists of a deep branch and a shallow branch.
The deep branch consists of convolutional layers with small kernel size focusing on details in textures, 
and the shallow branch consists of three convolutional layers with large kernel size focusing on larger-scale and non-local structures.
The combination of these two branches enables cgCNN to model both global and local structures.
When synthesizing dynamic or sound textures,
we use spatial-temporal or temporal convolutional layers respectively.
We use the hard sigmoid function as activation function in the network,
which is defined as:
\begin{equation}
\label{hardsigmoid}
\sigma(x) = \min(\max(x,0),1).
\end{equation}

\paragraph{Parameters} 
We sample $K=3$ textures in each iteration and each sample is initialized as Gaussian noise with variance $0.01$. 
We run $N=10$ or $50$ Langevin steps in each iteration.
The training algorithm stops with a maximal of $T=5000$ iterations. 
We use RMSprop~\cite{Tieleman2012} or Adam~\cite{kingma2014adam} to update networks and synthesized images, 
with the initial learning rate set to $0.001$. 
In all our experiments, 
we follow these settings except where explicitly stated. 

All the results are available at \url{http://captain.whu.edu.cn/cgcnn-texture/}, 
where one can check dynamic and sound textures.

\subsection{Bounded constraint}
\label{Bounded_experiment}
\begin{figure}[t!]
	\vspace{-2mm}
	\subfigure[Exemplar]{
		\begin{minipage}[b]{0.231\linewidth}
			\includegraphics[width=1\linewidth]{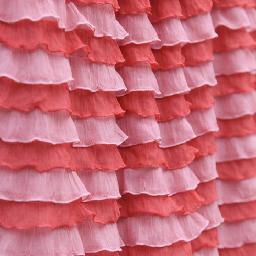} \\ \vspace{-2ex}
			\includegraphics[width=1\linewidth]{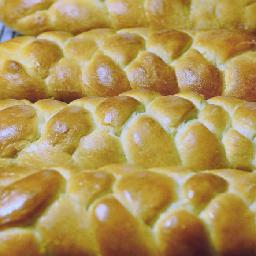}
		\end{minipage}
	}
	\hspace{-2ex}
	\subfigure[{\em hard sigmoid}]{
		\begin{minipage}[b]{0.231\linewidth}
			\includegraphics[width=1\linewidth]{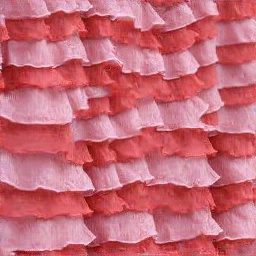}\\ \vspace{-2ex}
			\includegraphics[width=1\linewidth]{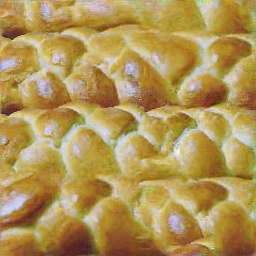}
		\end{minipage}
	}
	\hspace{-2ex}
	\subfigure[{\em tanh}]{
		\begin{minipage}[b]{0.231\linewidth}
			\includegraphics[width=1\linewidth]{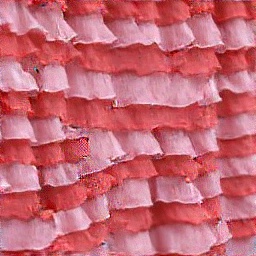}\\ \vspace{-2ex}
			\includegraphics[width=1\linewidth]{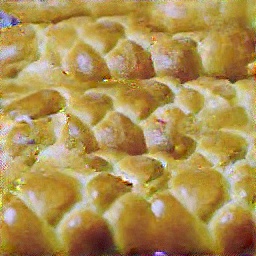}
		\end{minipage}
	}
	\hspace{-2ex}
	\subfigure[{\em sigmoid}]{
		\begin{minipage}[b]{0.231\linewidth}
			\includegraphics[width=1\linewidth]{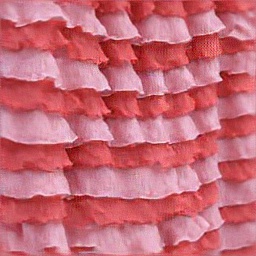} \\ \vspace{-2ex}
			\includegraphics[width=1\linewidth]{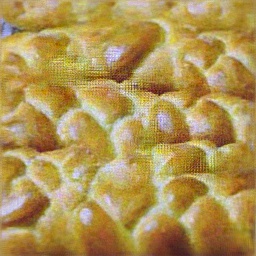}
		\end{minipage}
	}
\vspace{-4mm}
	\caption{Results using different activation functions. The {\em hard sigmoid} generates the most satisfactory results. Zooming in to check the artifacts generated by {\em tanh} and {\em sigmoid}.}
	\label{Compare_Activation}
\vspace{-3mm}
\end{figure}

We find it is crucial to constrain the magnitude of the energy $E_{cg}$ in order to stabilize the training process,
because the energy $E_{cg}$ often grows too large and causes the exploding gradient problem.
In this work,
we use bounded activation function in the network architecture to ensure the energy $E_{cg}$ is upper bounded.

We notice the choice of activation function has subtle influences on the synthesis results.
This is shown in Fig.~\ref{Compare_Activation}, 
where we present the results using different activation functions, 
\ie {\em hard sigmoid, tanh} and {\em sigmoid} respectively. 
We observe the use of {\em hard sigmoid} produces the most satisfactory results, 
while {\em tanh} often generates some unnatural colors, 
and the results using {\em sigmoid} exhibit some check-board artifacts.

Comparing with other constraints such as weight clipping~\cite{ArjovskyCB17}, 
gradient penalty~\cite{GulrajaniAADC17} and spectral normalization~\cite{abs-1802-05957},
our method does not have extra computational cost,
is nonparametric and easy to implement.

\subsection{Diversity of synthesis}
\label{Diversity_experiment}
\begin{figure}[t!]
	\vspace{-2mm}
		\subfigure[Exemplar]{
			\begin{minipage}[b]{0.231\linewidth}
				\includegraphics[width=1\linewidth]{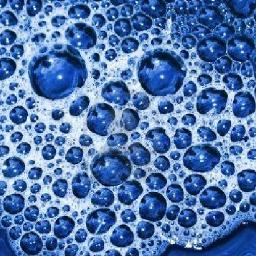}\\
				\vspace{-2ex}
				\includegraphics[width=1\linewidth]{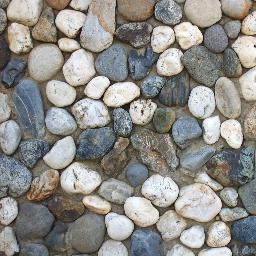}\\
				\vspace{-2ex}
			\end{minipage}
		}
		\subfigure[Our results]{
			\begin{minipage}[b]{0.77\linewidth}
				\begin{minipage}[b]{0.3\linewidth}
					\includegraphics[width=1\linewidth]{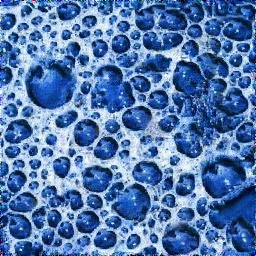}\\
					\vspace{-2ex}
					\includegraphics[width=1\linewidth]{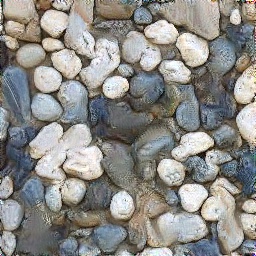}\\
					\vspace{-2ex}
				\end{minipage}
				\begin{minipage}[b]{0.3\linewidth}
					\includegraphics[width=1\linewidth]{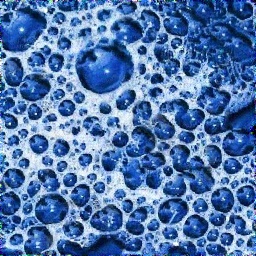}\\
					\vspace{-2ex}
					\includegraphics[width=1\linewidth]{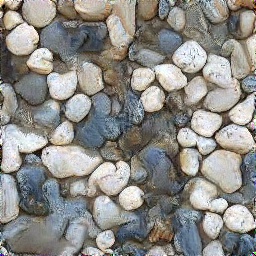}\\
					\vspace{-2ex}
				\end{minipage}
				\begin{minipage}[b]{0.3\linewidth}
					\includegraphics[width=1\linewidth]{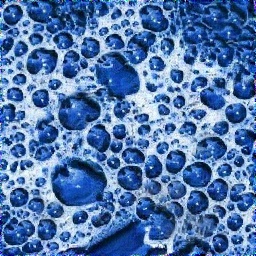}\\
					\vspace{-2ex}
					\includegraphics[width=1\linewidth]{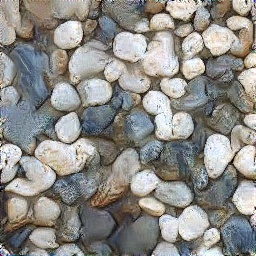}\\
					\vspace{-2ex}
				\end{minipage}
			\end{minipage}
		}
	\vspace{-4mm}
		\caption{Diversity of synthesis. The produced textures are visually similar to the inputs but are not identical to each other.}
		\label{Diversity}
		\vspace{-4mm}
\end{figure}

It is important for a texture synthesis algorithm to be able to synthesis diversified texture samples using a given exemplar.
For the proposed c-cgCNN model,
the diversity of the synthesized textures is a direct result of the randomness of the initial Gaussian noise, 
thus one does not need to make extra effort to ensure such diversity. 
This is shown in Fig.~\ref{Diversity},
where a batch of three synthesized samples for each exemplars are presented.
Note that all synthesized textures are visually similar to the exemplars, 
but they are not identical to each other.

\subsection{Ablation study of the learning algorithm}
In order to verify the importance of $\mathcal{D}$-learning step in Alg.~\ref{our_algorithm_1},
we test a fixed random method which is to disable $\mathcal{D}$-learning step.
This fixed random method is actually optimizing the synthesized image using a fixed random ConvNet.

\begin{figure}[h!]
	
	\vspace{-2mm}
		\begin{center}
		\subfigure[Exemplar]{
			\includegraphics[width=0.18\linewidth]{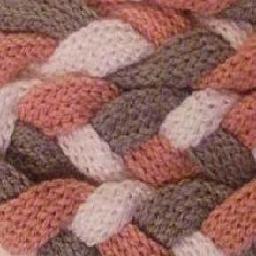}
		}
		\subfigure[Energy evolutions.]{
			\includegraphics[width=0.45\linewidth]{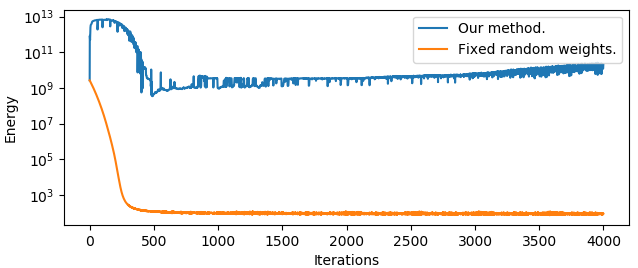}
		}\\
		\subfigure{
			\begin{minipage}[b]{0.03\linewidth}
				\rotatebox{90}{Ours} \\ \vspace{3mm}
				\rotatebox{90}{Random}
			\end{minipage}
		}
		\hspace{-1.3ex}
		\subfigure[Iter 0]{
			\begin{minipage}[b]{0.18\linewidth}
				\includegraphics[width=1\linewidth]{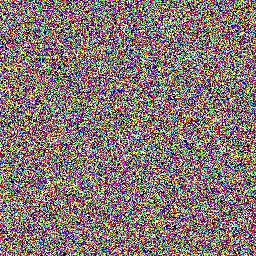} \\\vspace{-2ex}
				\includegraphics[width=1\linewidth]{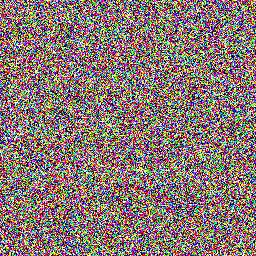} \\\vspace{-2ex}
			\end{minipage}
		}
		\subfigure[Iter 300]{
			\begin{minipage}[b]{0.18\linewidth}
				\includegraphics[width=1\linewidth]{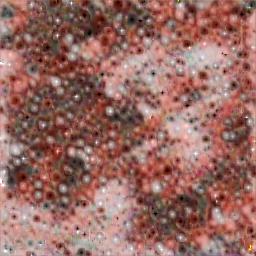} \\\vspace{-2ex}
				\includegraphics[width=1\linewidth]{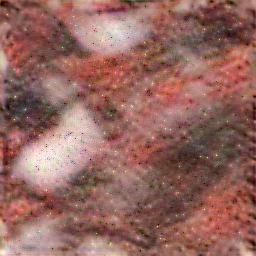} \\\vspace{-2ex}
			\end{minipage}
		}
		\subfigure[Iter 900]{
			\begin{minipage}[b]{0.18\linewidth}
				\includegraphics[width=1\linewidth]{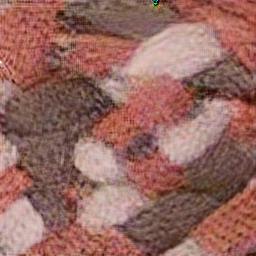} \\\vspace{-2ex}
				\includegraphics[width=1\linewidth]{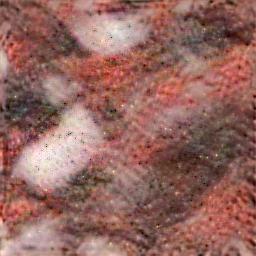} \\\vspace{-2ex}
			\end{minipage}
		}
		\subfigure[Iter 4000]{
			\begin{minipage}[b]{0.18\linewidth}
				\includegraphics[width=1\linewidth]{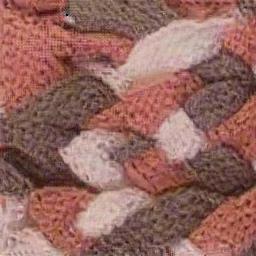} \\\vspace{-2ex}
				\includegraphics[width=1\linewidth]{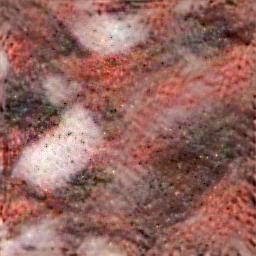} \\\vspace{-2ex}
			\end{minipage}
		}
		\end{center}
	 \vspace{-4mm}
		\caption{Compare with fixed random method. The differences highlight the effectiveness of $\mathcal{D}$-learning step in Alg.~\ref{our_algorithm_1}.}
		\label{random_optimization}
	 \vspace{-3mm}		
\end{figure}

Fig.~\ref{random_optimization} presents the comparison between our Alg.~\ref{our_algorithm_1} and such fixed random method.
Clearly, 
our method produces more favorable results than this fixed random method,
as our results are sharper and clearer while this method can only produce blurry and noisy textures.
We can therefore conclude that $\mathcal{D}$-learning step is key to the success of our algorithm,
as it enables us to learn better deep filters than a random ConvNet.

\subsection{Ablation study of the network architecture}
\label{Ablation_experiment}
\begin{figure*}[th!]
	\vspace{-2mm}
	\begin{center}
	\subfigure[Exemplar]{
		\begin{minipage}[b]{0.12\linewidth}
			\includegraphics[width=1\linewidth]{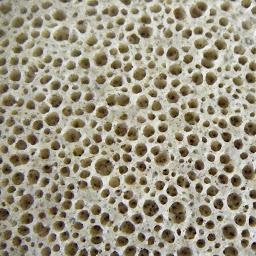}\\
			\vspace{-2ex}
			\includegraphics[width=1\linewidth]{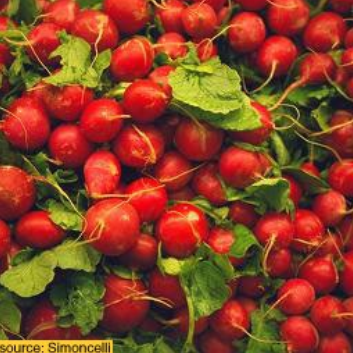}\\
			\vspace{-2ex}
			\includegraphics[width=1\linewidth]{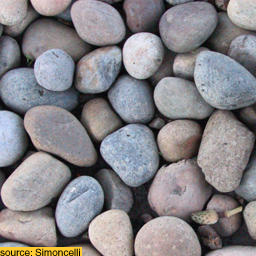}
		\end{minipage}
	}
	\hspace{-1.5ex}
	\subfigure[$(1 \mathbf D \oplus 0 \mathbf S)$]{
		\begin{minipage}[b]{0.12\linewidth}
			\includegraphics[width=1\linewidth]{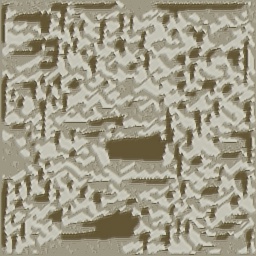}\\
			\vspace{-2ex}
			\includegraphics[width=1\linewidth]{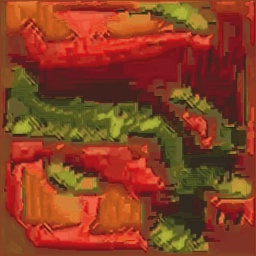}\\
			\vspace{-2ex}
			\includegraphics[width=1\linewidth]{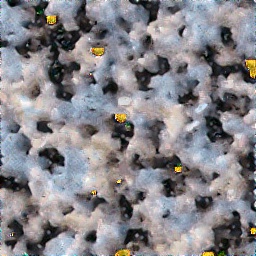}
		\end{minipage}
	}
	\hspace{-1.5ex}
	\subfigure[$(3 \mathbf D \oplus 0 \mathbf S)$]{
		\begin{minipage}[b]{0.12\linewidth}
			\includegraphics[width=1\linewidth]{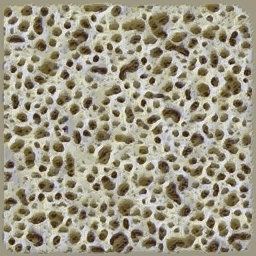}\\
			\vspace{-2ex}
			\includegraphics[width=1\linewidth]{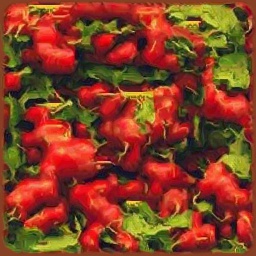}\\
			\vspace{-2ex}
			\includegraphics[width=1\linewidth]{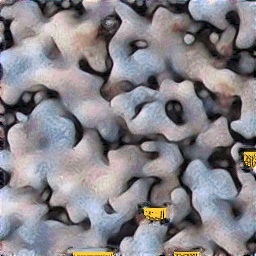}
		\end{minipage}
	}
	\hspace{-1.5ex}
	\subfigure[$(9 \mathbf D \oplus 0 \mathbf S)$]{
		\begin{minipage}[b]{0.12\linewidth}
			\includegraphics[width=1\linewidth]{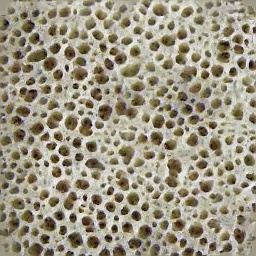}\\
			\vspace{-2ex}
			\includegraphics[width=1\linewidth]{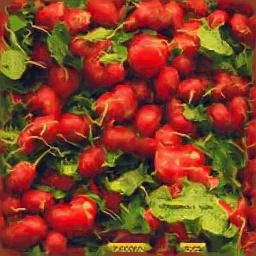}\\
			\vspace{-2ex}
			\includegraphics[width=1\linewidth]{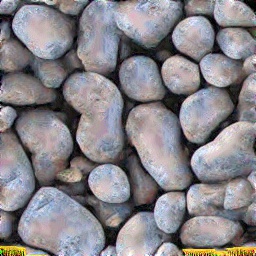}
		\end{minipage}
	}
	\hspace{-1.5ex}
	\subfigure[$(9 \mathbf D \oplus 1 \mathbf S)$]{
		\begin{minipage}[b]{0.12\linewidth}
			\includegraphics[width=1\linewidth]{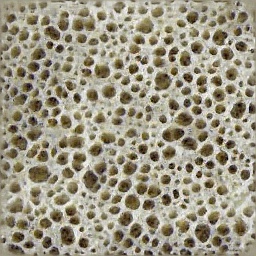}\\
			\vspace{-2ex}
			\includegraphics[width=1\linewidth]{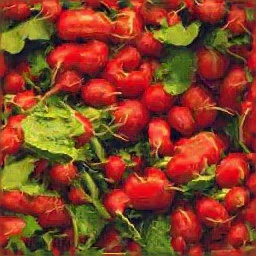}\\
			\vspace{-2ex}
			\includegraphics[width=1\linewidth]{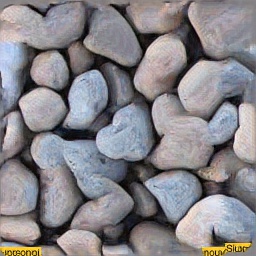}
		\end{minipage}
	}
	\hspace{-1.5ex}
	\subfigure[$(9 \mathbf D \oplus 3 \mathbf S)$]{
		\begin{minipage}[b]{0.12\linewidth}
			\includegraphics[width=1\linewidth]{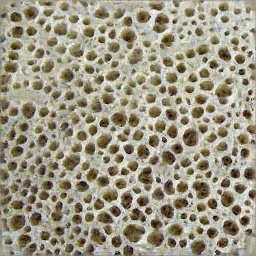}\\
			\vspace{-2ex}
			\includegraphics[width=1\linewidth]{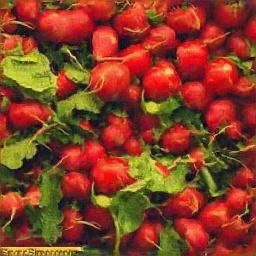}\\
			\vspace{-2ex}
			\includegraphics[width=1\linewidth]{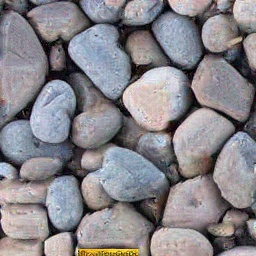}
		\end{minipage}
	}
	\end{center}
	\vspace{-4mm}
	\caption{Texture synthesized with different sub-network. One can check that the synthesized textures contain larger scale structures as the receptive field of the selected sub-network increases.}
	\label{Compare_Scale}
\end{figure*}

In order to investigate the roles played by different layers in our network in Fig.~\ref{network},
we carry out an ablation study by using different sub-networks.
Note the original network has two branches consisting of $9$ deep layers and $3$ shallow layers respectively.
We denote a sub-network with $m$ deep layers and $n$ shallow layers by $(m \mathbf D \oplus n \mathbf S)$. 
For instance, 
a sub-network $(3 \mathbf D \oplus 1 \mathbf S)$ consists of the first $3$ layers in the deep branch and the first $1$ layer in the shallow branch. 
We experiment with $m=1, 3, 9$ and $n=0, 1, 3$.
Fig.~\ref{Compare_Scale} presents the results of five sub-networks with increasingly large receptive field, 
\ie $(1 \mathbf D \oplus 0 \mathbf S), (3 \mathbf D \oplus 0 \mathbf S), (9\mathbf D \oplus 0\mathbf S), (9\mathbf D \oplus 1 \mathbf S), (9\mathbf D \oplus 3 \mathbf S)$.
As we can see, 
the synthesized textures capture larger scale structures as the receptive field increases. 

In general, 
to generate high fidelity samples, 
the network must be able to model structures of different scales contained in the input image. 
As shown in Fig.~\ref{Compare_Scale}, 
$(1 \mathbf D \oplus 0 \mathbf S)$ generates results with serious artifacts because the receptive field is only $3$ pixels wide,
which is too small for any meaningful texture elements.
For the {\em porous} texture which consists of small scale elements, 
a sub-network with a relatively small receptive field, 
e.g. $(3 \mathbf D \oplus 0 \mathbf S)$ and $(9\mathbf D \oplus 0\mathbf S)$, 
is sufficient to produce high quality textures. 
However, 
for textures containing larger-scale structures, 
like {\em cherries} and {\em pebbles}, 
larger receptive fields are often required for producing better results.

\subsection{Results on texture synthesis}
\label{Synthesis_experiment}
\begin{figure*}[h!]
		\begin{center}
		\subfigure[Exemplar]{
			\begin{minipage}[b]{0.13\linewidth}
				\includegraphics[width=1\linewidth]{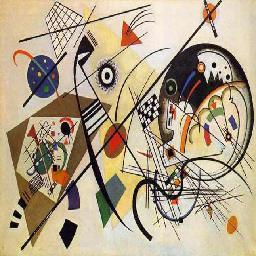}\\
				\vspace{-2ex}
				\includegraphics[width=1\linewidth]{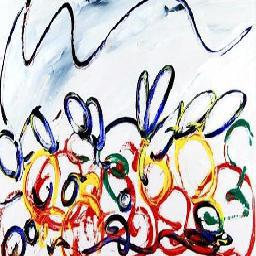}\\ \vspace{-2ex}
				\includegraphics[width=1\linewidth]{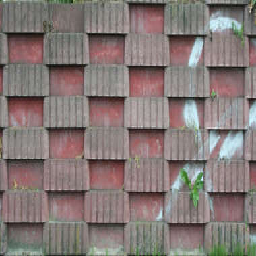} \\ \vspace{-2ex}
				\includegraphics[width=1\linewidth]{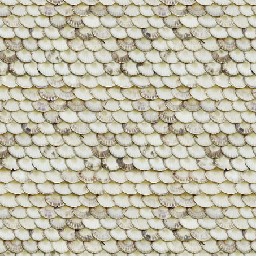} \\ \vspace{-2ex}
				\includegraphics[width=1\linewidth]{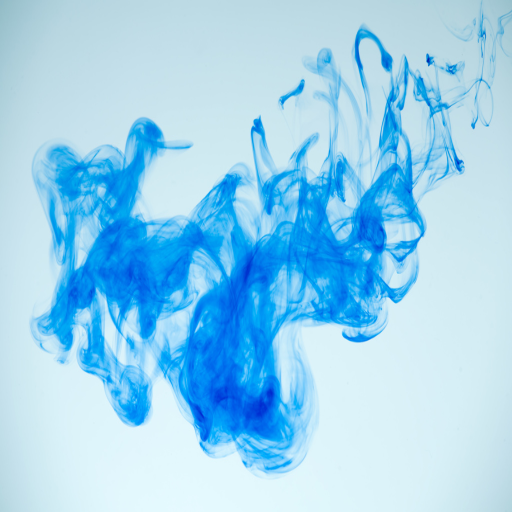} \\ \vspace{-2ex}
			\end{minipage}
		}
		\hspace{-1.5ex}
		\subfigure[Gatys'~\cite{gatys2015texture}]{
			\begin{minipage}[b]{0.13\linewidth}
				\includegraphics[width=1\linewidth]{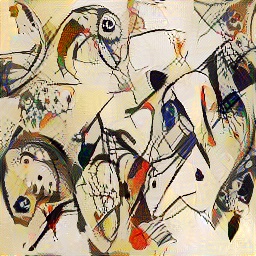} \\ \vspace{-2ex}
				\includegraphics[width=1\linewidth]{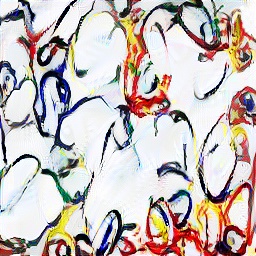} \\ \vspace{-2ex}
				\includegraphics[width=1\linewidth]{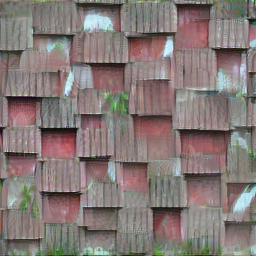} \\ \vspace{-2ex}
				\includegraphics[width=1\linewidth]{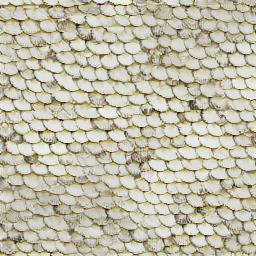} \\ \vspace{-2ex}
				\includegraphics[width=1\linewidth]{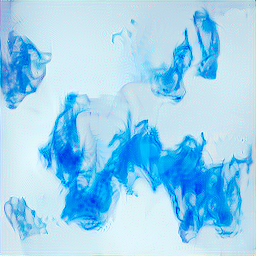} \\ \vspace{-2ex}
			\end{minipage}
		}
		\hspace{-1.5ex}
		\subfigure[gCNN~\cite{xie2016theory}]{
			\begin{minipage}[b]{0.13\linewidth}
				\includegraphics[width=1\linewidth]{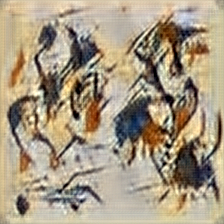} \\ \vspace{-2ex}
				\includegraphics[width=1\linewidth]{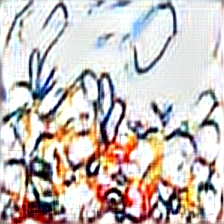} \\ \vspace{-2ex}
				\includegraphics[width=1\linewidth]{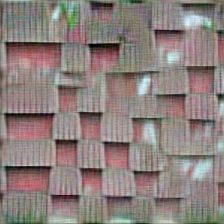} \\ \vspace{-2ex}
				\includegraphics[width=1\linewidth]{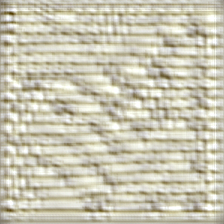} \\ \vspace{-2ex}
				\includegraphics[width=1\linewidth]{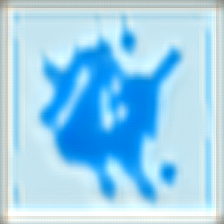}\\ \vspace{-2ex}
			\end{minipage}
		}
		\hspace{-1.5ex}
		\subfigure[CoopNet~\cite{xie2018cooperative}]{
			\begin{minipage}[b]{0.13\linewidth}
				\includegraphics[width=1\linewidth]{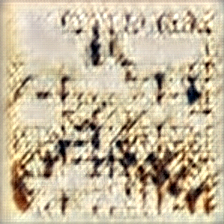} \\ \vspace{-2ex}
				\includegraphics[width=1\linewidth]{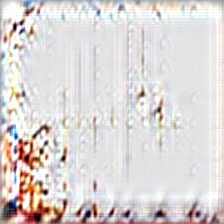} \\ \vspace{-2ex}
				\includegraphics[width=1\linewidth]{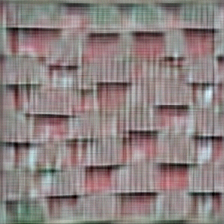} \\ \vspace{-2ex}
				\includegraphics[width=1\linewidth]{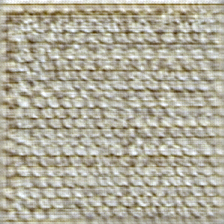} \\ \vspace{-2ex}
				\includegraphics[width=1\linewidth]{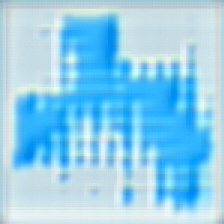}\\ \vspace{-2ex}
			\end{minipage}
		}
		\hspace{-1.5ex}
		\subfigure[Self tuning~\cite{kaspar2015self}]{
			\begin{minipage}[b]{0.13\linewidth}
				\includegraphics[width=1\linewidth]{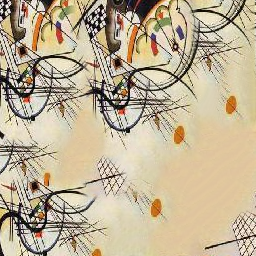}\\ \vspace{-2ex}
				\includegraphics[width=1\linewidth]{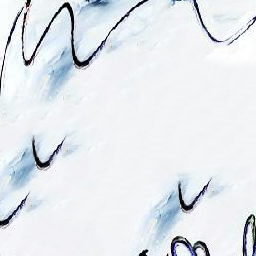}\\ \vspace{-2ex}
				\includegraphics[width=1\linewidth]{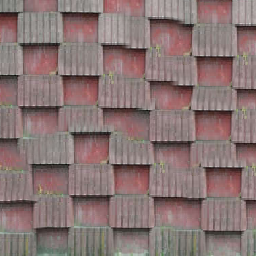}\\ \vspace{-2ex}
				\includegraphics[width=1\linewidth]{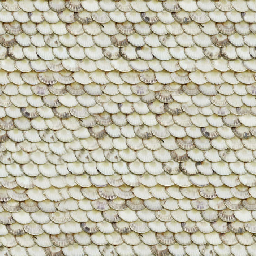}\\ \vspace{-2ex}
				\includegraphics[width=1\linewidth]{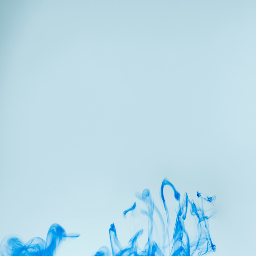}\\ \vspace{-2ex}
			\end{minipage}
		}
		\hspace{-1.5ex}
		\subfigure[c-cgCNN-Mean]{
			\begin{minipage}[b]{0.13\linewidth}
				\includegraphics[width=1\linewidth]{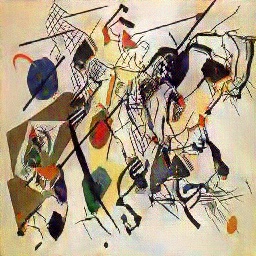} \\ \vspace{-2ex}
				\includegraphics[width=1\linewidth]{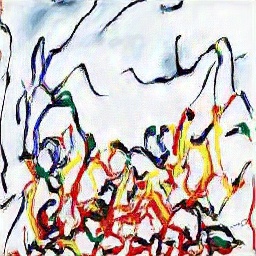} \\ \vspace{-2ex}
				\includegraphics[width=1\linewidth]{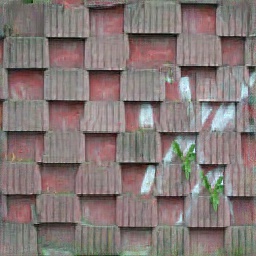} \\ \vspace{-2ex}
				\includegraphics[width=1\linewidth]{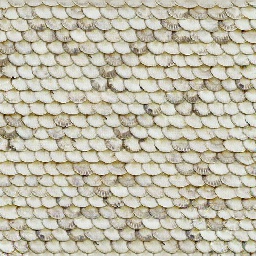} \\ \vspace{-2ex}
				\includegraphics[width=1\linewidth]{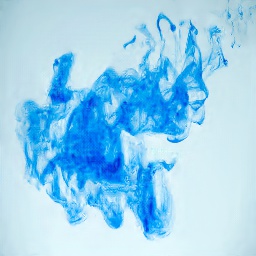} \\ \vspace{-2ex}
			\end{minipage}
		}
		\hspace{-1.5ex}
		\subfigure[c-cgCNN-Gram]{
			\begin{minipage}[b]{0.13\linewidth}
				\includegraphics[width=1\linewidth]{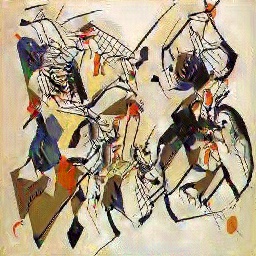} \\ \vspace{-2ex}
				\includegraphics[width=1\linewidth]{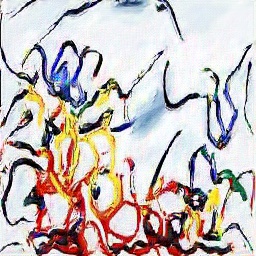} \\ \vspace{-2ex}
				\includegraphics[width=1\linewidth]{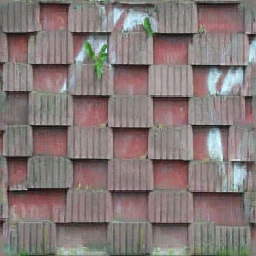} \\ \vspace{-2ex}
				\includegraphics[width=1\linewidth]{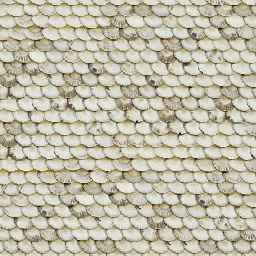} \\ \vspace{-2ex}
				\includegraphics[width=1\linewidth]{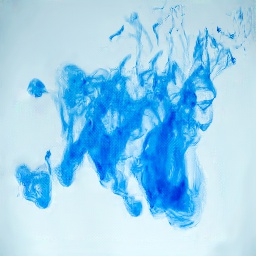} \\ \vspace{-2ex}
			\end{minipage}
		}
		\end{center} 
	\vspace{-4mm}
		\caption{Textures synthesized by different methods. See texts for more details.}
		\label{Compare_All}
		\vspace{-4mm}
\end{figure*}

\begin{figure*}[t!]
	
	\vspace{-4mm}
		\begin{center}
			\includegraphics[width=0.1\linewidth]{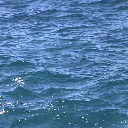}
			\includegraphics[width=0.1\linewidth]{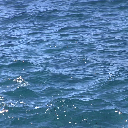}
			\includegraphics[width=0.1\linewidth]{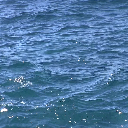}
			\includegraphics[width=0.1\linewidth]{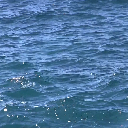}
			\includegraphics[width=0.1\linewidth]{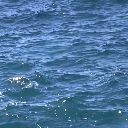}
			\includegraphics[width=0.1\linewidth]{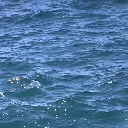}
			\includegraphics[width=0.1\linewidth]{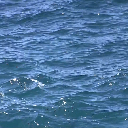}
			\includegraphics[width=0.1\linewidth]{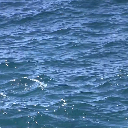}
			\includegraphics[width=0.1\linewidth]{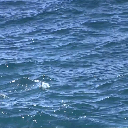}\\
			\vspace{1mm}

			\includegraphics[width=0.1\linewidth]{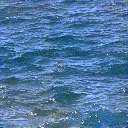}
			\includegraphics[width=0.1\linewidth]{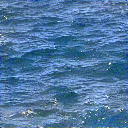}
			\includegraphics[width=0.1\linewidth]{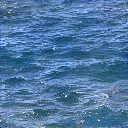}
			\includegraphics[width=0.1\linewidth]{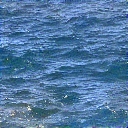}
			\includegraphics[width=0.1\linewidth]{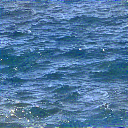}
			\includegraphics[width=0.1\linewidth]{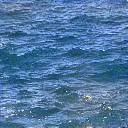}
			\includegraphics[width=0.1\linewidth]{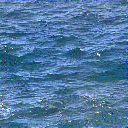}
			\includegraphics[width=0.1\linewidth]{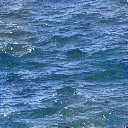}
			\includegraphics[width=0.1\linewidth]{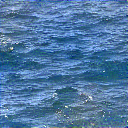} \\
			\vspace{1mm}

			\includegraphics[width=0.1\linewidth]{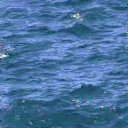}
			\includegraphics[width=0.1\linewidth]{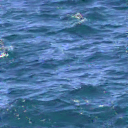}
			\includegraphics[width=0.1\linewidth]{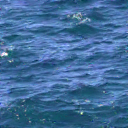}
			\includegraphics[width=0.1\linewidth]{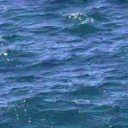}
			\includegraphics[width=0.1\linewidth]{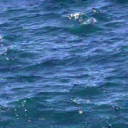}
			\includegraphics[width=0.1\linewidth]{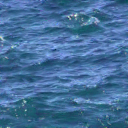}
			\includegraphics[width=0.1\linewidth]{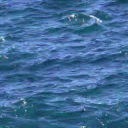}
			\includegraphics[width=0.1\linewidth]{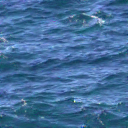}
			\includegraphics[width=0.1\linewidth]{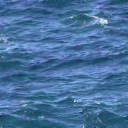} \\
			\vspace{1mm}

			\includegraphics[width=0.1\linewidth]{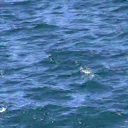}
			\includegraphics[width=0.1\linewidth]{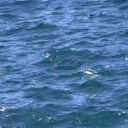}
			\includegraphics[width=0.1\linewidth]{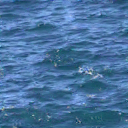}
			\includegraphics[width=0.1\linewidth]{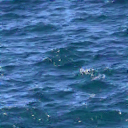}
			\includegraphics[width=0.1\linewidth]{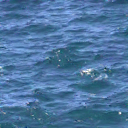}
			\includegraphics[width=0.1\linewidth]{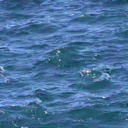}
			\includegraphics[width=0.1\linewidth]{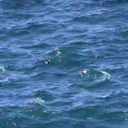}
			\includegraphics[width=0.1\linewidth]{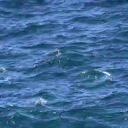}
			\includegraphics[width=0.1\linewidth]{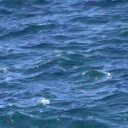} \\
			\vspace{1mm}

			\includegraphics[width=0.1\linewidth]{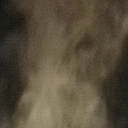}
			\includegraphics[width=0.1\linewidth]{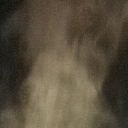}
			\includegraphics[width=0.1\linewidth]{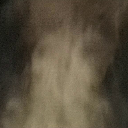}
			\includegraphics[width=0.1\linewidth]{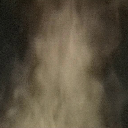}
			\includegraphics[width=0.1\linewidth]{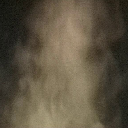}
			\includegraphics[width=0.1\linewidth]{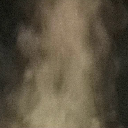}
			\includegraphics[width=0.1\linewidth]{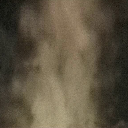}
			\includegraphics[width=0.1\linewidth]{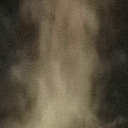}
			\includegraphics[width=0.1\linewidth]{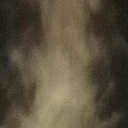}\\
			\vspace{1mm}

			\includegraphics[width=0.1\linewidth]{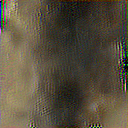}
			\includegraphics[width=0.1\linewidth]{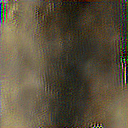}
			\includegraphics[width=0.1\linewidth]{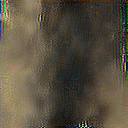}
			\includegraphics[width=0.1\linewidth]{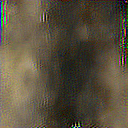}
			\includegraphics[width=0.1\linewidth]{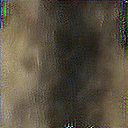}
			\includegraphics[width=0.1\linewidth]{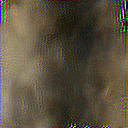}
			\includegraphics[width=0.1\linewidth]{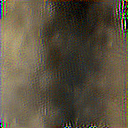}
			\includegraphics[width=0.1\linewidth]{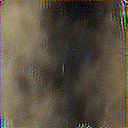}
			\includegraphics[width=0.1\linewidth]{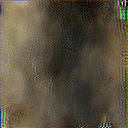} \\
			\vspace{1mm}

			\includegraphics[width=0.1\linewidth]{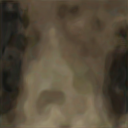}
			\includegraphics[width=0.1\linewidth]{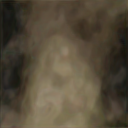}
			\includegraphics[width=0.1\linewidth]{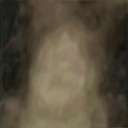}
			\includegraphics[width=0.1\linewidth]{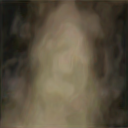}
			\includegraphics[width=0.1\linewidth]{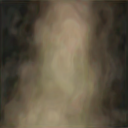}
			\includegraphics[width=0.1\linewidth]{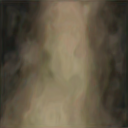}
			\includegraphics[width=0.1\linewidth]{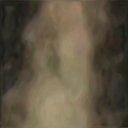}
			\includegraphics[width=0.1\linewidth]{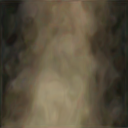}
			\includegraphics[width=0.1\linewidth]{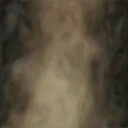} \\
			\vspace{1mm}

			\includegraphics[width=0.1\linewidth]{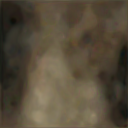}
			\includegraphics[width=0.1\linewidth]{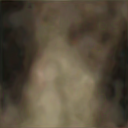}
			\includegraphics[width=0.1\linewidth]{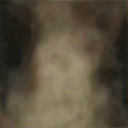}
			\includegraphics[width=0.1\linewidth]{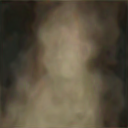}
			\includegraphics[width=0.1\linewidth]{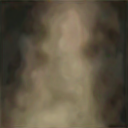}
			\includegraphics[width=0.1\linewidth]{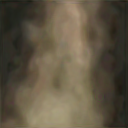}
			\includegraphics[width=0.1\linewidth]{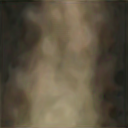}
			\includegraphics[width=0.1\linewidth]{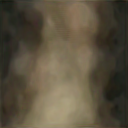}
			\includegraphics[width=0.1\linewidth]{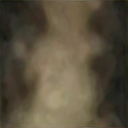} \\
		\end{center}
	\vspace{-4mm}
		\caption{Comparison between c-cgCNN and two-stream algorithm~\cite{tesfaldet2018two} for dynamic texture synthesis.
		For each dynamic texture, we present the exemplar (1-st row), results of two-stream method (2-nd row), c-cgCNN-Gram (3-rd row) and c-cgCNN-Mean (4-th row).
		While two-stream method suffers from low level noise and greyish effect, our method is free from these artifacts.} 
		\label{Dynamic_texture}
\end{figure*}

\begin{figure*}[htb!]
	\vspace{-2mm}
		\begin{center}
			\subfigure[Exemplar]{
				\begin{minipage}[b]{0.18\linewidth}
					\includegraphics[width=1\linewidth]{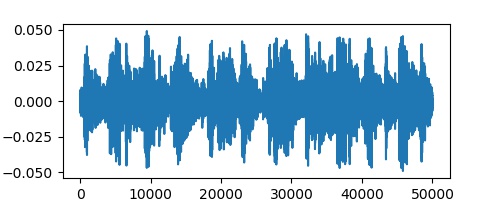}
				\end{minipage}
			}
			\subfigure[McDermott's~\cite{mcdermott2009sound}]{
				\begin{minipage}[b]{0.18\linewidth}
					\includegraphics[width=1\linewidth]{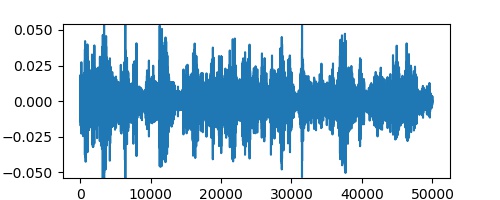}
				\end{minipage}
			}
			\subfigure[Antognini's~\cite{antognini2019audio}]{
				\begin{minipage}[b]{0.18\linewidth}
					\includegraphics[width=1\linewidth]{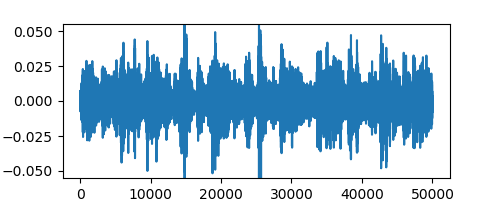}
				\end{minipage}
			}
			\subfigure[c-cgCNN-Gram]{
				\begin{minipage}[b]{0.18\linewidth}
					\includegraphics[width=1\linewidth]{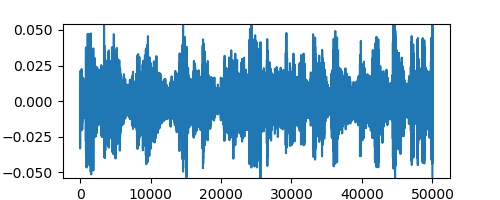}
				\end{minipage}
			}
			\subfigure[c-cgCNN-Mean]{
				\begin{minipage}[b]{0.18\linewidth}
					\includegraphics[width=1\linewidth]{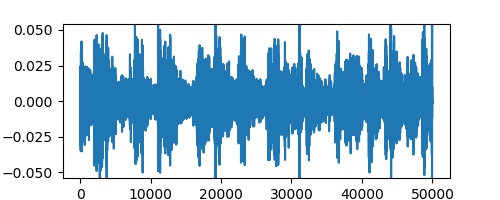}
				\end{minipage}
			}
		\end{center}
	\vspace{-3mm}
		\caption{Comparison of c-cgCNN, McDermott's~\cite{mcdermott2009sound} and Antognini's model~\cite{antognini2019audio} for sound texture synthesis. Their results are comparable. Sound texture shown here is ``applause''.}
		\label{Sound_texture}
		\vspace{-4mm}
\end{figure*}

For image texture synthesis,
we compare the following methods, 
which are theoretically related to our model or reported state-of-the-art performance. 
\begin{itemize}
\item[-] {\textbf{c-cgCNN-Gram}}: Our c-cgCNN with the Gram matrix as the statistic measurement.
\item[-] {\textbf{c-cgCNN-Mean}}: Our c-cgCNN where the mean vector is used as the statistic measurement instead of Gram matrix.
\item[-] {\textbf{Gatys' method}~\cite{gatys2015texture}}: A texture model relying on pretrained ConvNets. It is a special case of our \textbf{c-cgCNN-Gram} model with pretrained ConvNet.
\item[-] {\textbf{gCNN}~\cite{xie2016theory}}: A generative model reviewed in Sec.~\ref{Preliminaries}. It is a variation of our \textbf{c-cgCNN} model without deep statistics. 
\item[-] {\textbf{CoopNet}~\cite{xie2018cooperative}}: A generative model reviewed in Sec.~\ref{Preliminaries}. It is a combination of gCNN and a latent variable model. 
\item[-] {\textbf{Self tuning}~\cite{kaspar2015self}}: A recent patch-based EBTS algorithm that utilizes optimization technique.
\end{itemize}

Fig.~\ref{Compare_All} shows the qualitative comparison of these algorithms. 
We observe that Gatys' method fails to capture global structures (the 3-rd and 4-th textures) because the optimization process converges to a local minimum where global structures are not preserved, 
and it also generates artifacts such as unnatural color and noises (Zoom in the 1-st and 5-th textures). 
Meanwhile,
although gCNN and CoopNet can capture most of the large-scale structures, 
they loss too many details in the results, 
probably because they do not use any deep statistics.
Self tuning is excel at generating regular textures (the 3-rd and 4-th textures), 
but it sometimes losses the global structures (the 1-st, 2-nd and 4-th textures) due to the lack of global structure modeling in this method. 
In contrast,
c-cgCNN-Gram and c-cgCNN-Mean can both produce better samples than other baseline methods,
since they not only capture large-scale structures but also reproduce small-scale details,
even for highly structured textures (1-st, 3-rd and 4-th textures).
This is because c-cgCNN use both deep statistics and effective sampling strategy that are not likely to be trapped in bad local minimums.
It is also worth noticing that the results of c-cgCNN-Gram and c-cgCNN-Mean are comparable in most cases even though they use different statistics.
For quantitative evaluation,
we measure multi-scale structural similarity~\cite{Multiscale} (MS-SSIM) between the synthesized texture and the exemplar.
A higher score indicate higher visual similarity.
The quantitative results are summarized in Tab.~\ref{model_comparison_quantitative}.
The results show that our methods outperform other baseline methods in most cases.

\begin{table}[h!]
	\scriptsize
	\begin{center}
	  \caption{Quantitative evaluation of texture synthesis results shown in Fig.~\ref{Compare_All} using MS-SSIM.}
	  \label{model_comparison_quantitative}
	  \begin{tabular}{r c c c c c c} 
		\hline
		& painting & lines & wall & scaly & ink \\
		\hline
		Gatys' ~\cite{gatys2015texture}     &0.01 & 0.01& 0.09& 0.08&  0.34 \\
		gCNN~\cite{xie2016theory}           &0.05 & 0.05& 0.11& 0.11&  0.35 \\
		self-tuning~\cite{kaspar2015self}  &0.03 & \textbf{0.17}& 0.07& 0.01&  0.42 \\
		CoopNet~\cite{xie2018cooperative}   &0.05 & 0.09& 0.20& 0.08&  0.32 \\
		\hline
		c-cgCNN-Gram        &0.10 & 0.09& \textbf{0.31}& \textbf{0.36} &  0.43\\
		c-cgCNN-Mean        &\textbf{0.14} & 0.10& 0.10& 0.00&  \textbf{0.46} \\
		\hline
	  \end{tabular}
	\end{center}
	\vspace{-4mm}
\end{table}

For dynamic texture synthesis,
we use the network $(6 \mathbf D \oplus 0 \mathbf S)$,
and we sample $M=2$ dynamic textures in sampling step. 
Fig.~\ref{Dynamic_texture} presents the qualitative comparison between c-cgCNN method and recent advanced two-stream method~\cite{tesfaldet2018two}.
We notice the results of two-stream model suffer from artifacts such as greyish (the 1-st texture) or low level noise (the 2-nd texture),
and sometime exhibit temporal inconsistency.
While the results of both c-cgCNN-Gram and c-cgCNN-Mean are more favorable as they are cleaner and show better temporal consistency.

For qualitative evaluation,
we measure the average of MS-SSIM metric between each frame of synthesized results and the corresponding frame in the exemplar.
The results are shown in Tab.~\ref{Dynamic_texture_quantitative},
where both c-cgCNN-Mean and c-cgCNN-Gram outperform two-stream method.

\begin{table}[h!]
	\scriptsize
	\begin{center}
	  \caption{Quantitative evaluation of dynamic texture synthesis results shown in Fig.~\ref{Dynamic_texture} using MS-SSIM.}
	  \label{Dynamic_texture_quantitative}
	  \begin{tabular}{r c c} 
		\hline
		&  ocean & smoke \\
		\hline
		TwoStream~\cite{tesfaldet2018two}  & 0.08 & 0.01  \\
		\hline
		c-cgCNN-Gram & \textbf{0.17} & 0.79  \\
		c-cgCNN-Mean & 0.13  &  \textbf{0.86}   \\
		\hline
	  \end{tabular}
	\end{center}
	\vspace{-4mm}
\end{table}

For sound texture synthesis,
we use the network $(4 \mathbf D \oplus 0 \mathbf S)$ where the kernel size and number of filters in each layer are $25$ and $128$,
and the strides in each layer is $10$ except the first layer where the stride is $5$.
We do not use pooling layers in this network.

Fig.~\ref{Sound_texture} presents the results of sound texture synthesis using c-cgCNN, McDermott's model~\cite{mcdermott2009sound} and Antognini's model~\cite{antognini2019audio} in waveforms.
Unlike other two methods which act on frequency domain,
c-cgCNN only uses raw audios.
We observe that the results of these methods are generally comparable,
except for some cases where our results are noisier than baseline methods.
It is probably because of the loss of short temporal dependencies caused by the large strides in the shallow layers.
It also suggests that our results might be further improved by using more carefully designed networks.

\subsection{Results on texture expansion}
\label{Expansion_experiment}
The structure of generator net $\mathcal{G}_{\theta}$ used in f-cgCNN is borrowed from TextureNet,
with two extra residual blocks at the output layer.
See~\cite{ulyanov2017improved} for details.
When expanding dynamic or sound texture,
the spatial convolutional layers in $\mathcal{G}_{\theta}$ is replaced by spatial-temporal or temporal convolutional layers accordingly.

\begin{figure}[htb!]
		\begin{center}
			\subfigure{
				\includegraphics[width=0.25\linewidth]{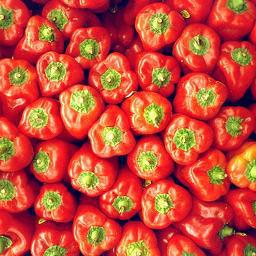}
				\includegraphics[width=0.37\linewidth]{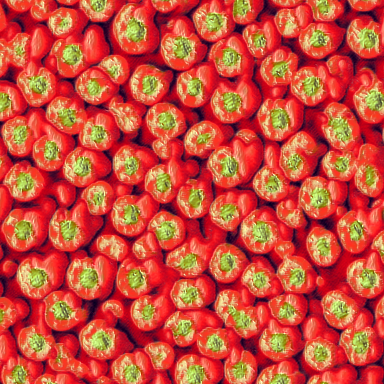}
				\includegraphics[width=0.37\linewidth]{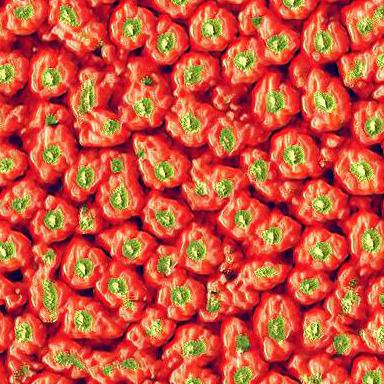}
			}\\
			\subfigure{
				\includegraphics[width=0.25\linewidth]{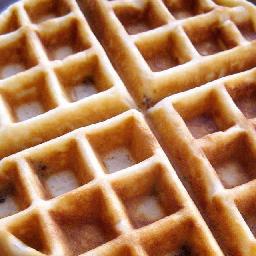}	
				\includegraphics[width=0.37\linewidth]{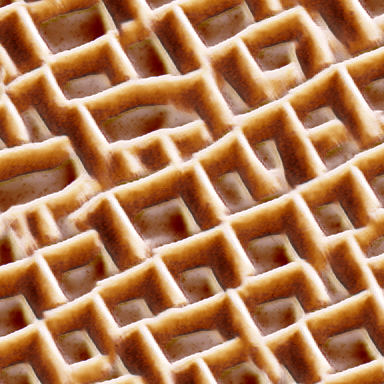}
				\includegraphics[width=0.37\linewidth]{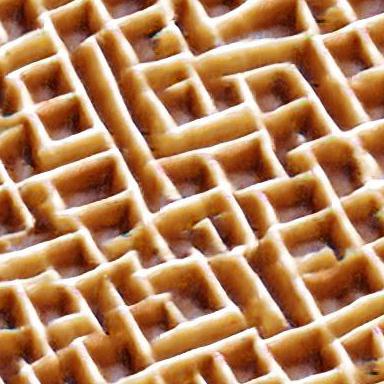}
			}\\
		\end{center}
	\vspace{-4mm}
		\caption{Comparison between TextureNet (2-nd column) and f-cgCNN (3-rd column) for image texture expansion. Their results are comparable.}
		\label{Compare_TextureNet}
		\vspace{-2mm}
\end{figure}

Fig.~\ref{Compare_TextureNet} presents a comparison between f-cgCNN and TextureNet in image texture expansion.
The results of f-cgCNN and TextureNet are generally comparable,
because both of them are able to learn the stationary elements in the exemplars.
In addition,
f-cgCNN are generally slower to converge than TextureNet in the training phase because it trains an extra net $\mathcal{D}$,
but their synthesis speed is the same as their synthesis both involve a forward pass through the generator net.

\begin{figure}[htb!]
	
		\begin{flushleft}
			\includegraphics[width=0.15\linewidth]{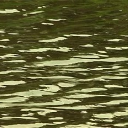}
			\includegraphics[width=0.15\linewidth]{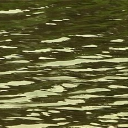}
			\includegraphics[width=0.15\linewidth]{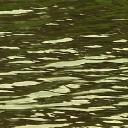}
			\includegraphics[width=0.15\linewidth]{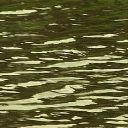} \\
			\vspace{1mm}

			\includegraphics[width=0.23\linewidth]{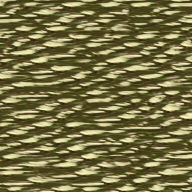}
			\includegraphics[width=0.23\linewidth]{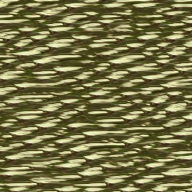}
			\includegraphics[width=0.23\linewidth]{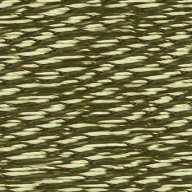}
			\includegraphics[width=0.23\linewidth]{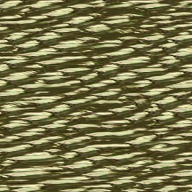} \\
			\vspace{1mm}

		\end{flushleft}
	\vspace{-4mm}   
		\caption{Results of dynamic texture expansion using f-cgCNN. We present the first 4 frames of exemplar (1-st row) and the first 4 frames of the synthesized results (2-nd row).}
		\label{Dynamic_Expansion}
		\vspace{-2mm}
\end{figure}

Fig.~\ref{Dynamic_Expansion} presents the results of dynamic texture expansion using f-cgCNN.
The exemplar dynamic texture is expanded to 48 frames and the size of each frame is $512 \times 512$.
We observe that f-cgCNN successfully reproduces stationary elements and expands the exemplar dynamic texture in both temporal and spatial dimensions,
\ie the synthesized textures have more frames and each frame is larger than the input exemplars.
It should be noticed that f-cgCNN is the first neural texture model that enables us to expand dynamic textures.

\begin{figure}[htb!]
	\vspace{-2mm}
		\begin{center}
			\subfigure[Exemplar]{
					\includegraphics[width=0.68\linewidth]{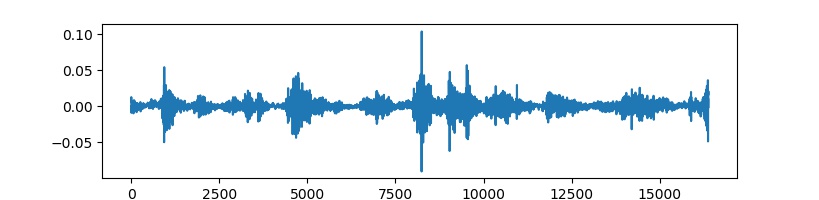}
			}\\
			\vspace{-4mm}  
			\subfigure[f-cgCNN]{
					\includegraphics[width=1\linewidth]{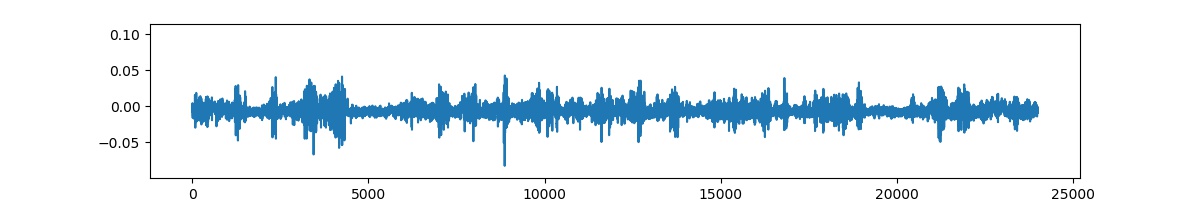}
			}
		\end{center}
	\vspace{-4mm}  
		\caption{Results of sound texture expansion using f-cgCNN. Sound texture shown here is ``shaking paper''.}
		\label{Sound_Expansion}
	\vspace{-4mm}
\end{figure}

Fig.~\ref{Sound_Expansion} presents the results of sound texture expansion.
In this experiment, 
we clip the first 16384 data points (less than 1 second) in each sound texture as exemplars,
and expand the exemplar to 122880 data points (about 5 seconds) using f-cgCNN.
Similar to the case of dynamic texture expansion,
f-cgCNN successfully expands the exemplar sound texture while preserving sound elements that occur most frequently.
Notice f-cgCNN is also the first texture model that enables us to expand sound textures.

\subsection{Results on texture inpainting}
\label{Inpainting_experiment}

\begin{figure}[htb!]
		\subfigure[Corrupted]{
			\begin{minipage}[b]{0.231\linewidth}
				\includegraphics[width=1\linewidth]{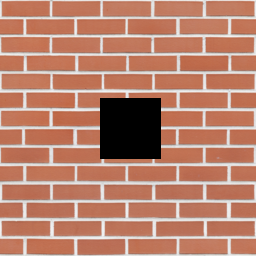} \\
				\vspace{-2ex}
				\includegraphics[width=1\linewidth]{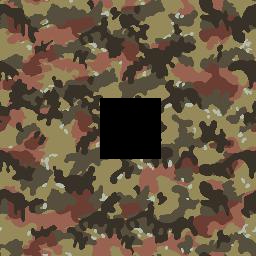} \\
				\vspace{-2ex}
				\includegraphics[width=1\linewidth]{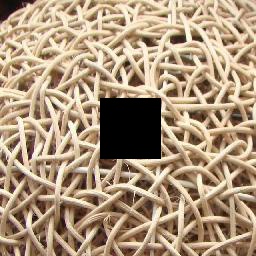} \\
				\vspace{-2ex}
				\includegraphics[width=1\linewidth]{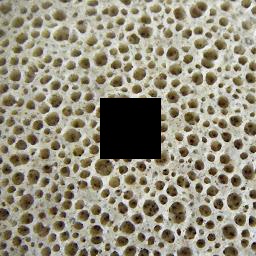} \\
				\vspace{-2ex}
				\includegraphics[width=1\linewidth]{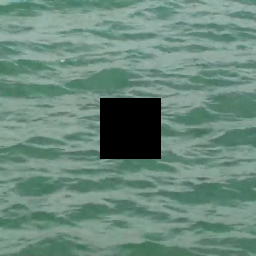} 
			\end{minipage}
		}
		\hspace{-2ex}
		\subfigure[Deep prior~\cite{ulyanov2018deep}]{
			\begin{minipage}[b]{0.231\linewidth}
				\includegraphics[width=1\linewidth]{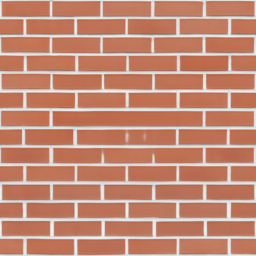} \\
				\vspace{-2ex}
				\includegraphics[width=1\linewidth]{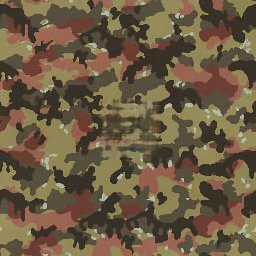} \\
				\vspace{-2ex}
				\includegraphics[width=1\linewidth]{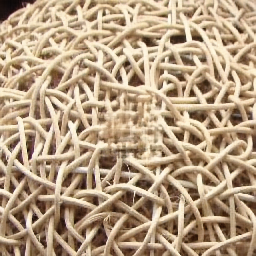} \\
				\vspace{-2ex}
				\includegraphics[width=1\linewidth]{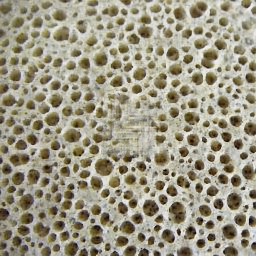} \\
				\vspace{-2ex}
				\includegraphics[width=1\linewidth]{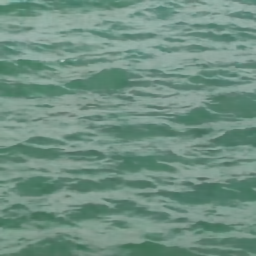}
			\end{minipage}
		}
		\hspace{-2ex}
		\subfigure[Deep fill~\cite{yu2018generative}]{
			\begin{minipage}[b]{0.231\linewidth}
				\includegraphics[width=1\linewidth]{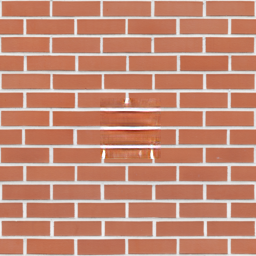} \\
				\vspace{-2ex}
				\includegraphics[width=1\linewidth]{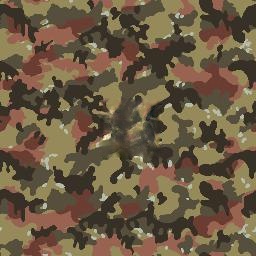} \\
				\vspace{-2ex}
				\includegraphics[width=1\linewidth]{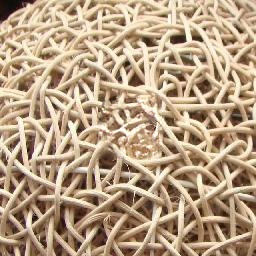} \\
				\vspace{-2ex}
				\includegraphics[width=1\linewidth]{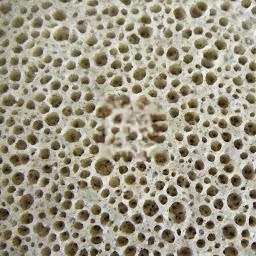} \\
				\vspace{-2ex}
				\includegraphics[width=1\linewidth]{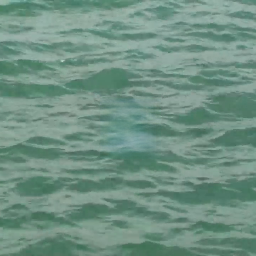} 
			\end{minipage}
		}
		\hspace{-2ex}
		\subfigure[cgCNN]{
			\begin{minipage}[b]{0.231\linewidth}
				\includegraphics[width=1\linewidth]{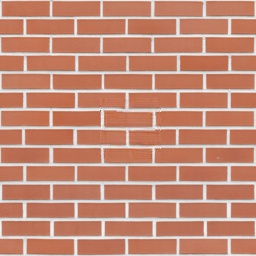}\\
				\vspace{-2ex}
				\includegraphics[width=1\linewidth]{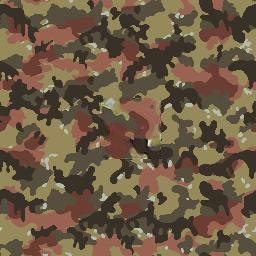}\\
				\vspace{-2ex}
				\includegraphics[width=1\linewidth]{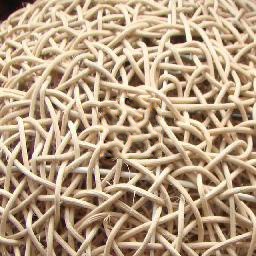}\\
				\vspace{-2ex}
				\includegraphics[width=1\linewidth]{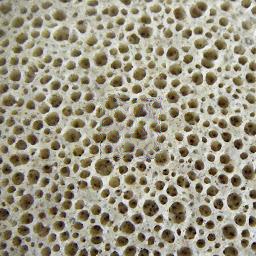}\\
				\vspace{-2ex}
				\includegraphics[width=1\linewidth]{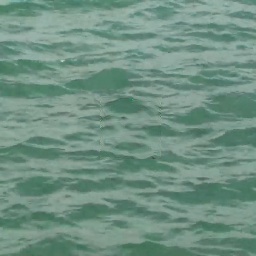} 
			\end{minipage}
		}
		\hspace{-2ex}
		\caption{Comparison of several neural inpainting methods. 
		It can be seen that our method produces the clearest results, 
		while the results of other methods are relatively blurry.}
		\label{image_inpainting}
		\vspace{-4mm}
\end{figure}

\begin{table}[b!]
	\scriptsize
	\begin{center}
	  \caption{Quantitative evaluation of inpainting results shown in Fig.~\ref{image_inpainting} using MS-SSIM.}
	  \label{image_inpainting_quantitative}
	  \begin{tabular}{r c c c c c c c} 
		\hline
		& brick & camouflage & fiber & sponge & water \\
		\hline
		DeepPrior~\cite{ulyanov2018deep} &0.978 & 0.897 & 0.856 & 0.904  & 0.956 \\
		DeepFill~\cite{yu2018generative} &0.966 & 0.922 & 0.900 & 0.900 & \textbf{0.962} \\
		\hline
		cgCNN &\textbf{0.984} & \textbf{0.930} & \textbf{0.905} & \textbf{0.914} & 0.912 \\
		\hline
	  \end{tabular}
	\end{center}
	\vspace{-4mm}
\end{table}

\begin{figure*}[bt!]
	
	\vspace{-2mm}
		\begin{center}
		\subfigure{\includegraphics[width=0.09\linewidth]{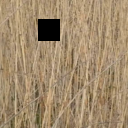}}
		\subfigure{\includegraphics[width=0.09\linewidth]{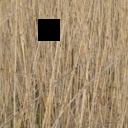}}
		\subfigure{\includegraphics[width=0.09\linewidth]{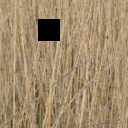}}
		\subfigure{\includegraphics[width=0.09\linewidth]{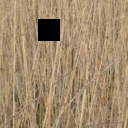}}
		\subfigure{\includegraphics[width=0.09\linewidth]{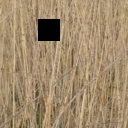}}
		\subfigure{\includegraphics[width=0.09\linewidth]{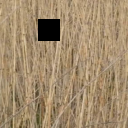}}
		\subfigure{\includegraphics[width=0.09\linewidth]{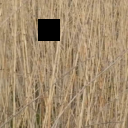}}
		\subfigure{\includegraphics[width=0.09\linewidth]{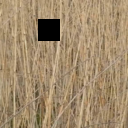}}
		\subfigure{\includegraphics[width=0.09\linewidth]{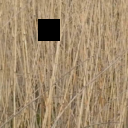}}
		\subfigure{\includegraphics[width=0.09\linewidth]{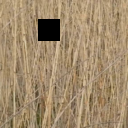}}\\
		\vspace{-1.5ex}

		\subfigure{\includegraphics[width=0.09\linewidth]{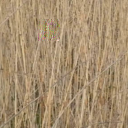}}
		\subfigure{\includegraphics[width=0.09\linewidth]{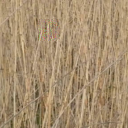}}
		\subfigure{\includegraphics[width=0.09\linewidth]{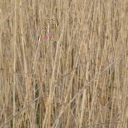}}
		\subfigure{\includegraphics[width=0.09\linewidth]{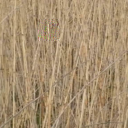}}
		\subfigure{\includegraphics[width=0.09\linewidth]{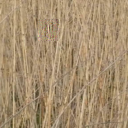}}
		\subfigure{\includegraphics[width=0.09\linewidth]{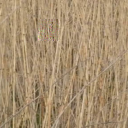}}
		\subfigure{\includegraphics[width=0.09\linewidth]{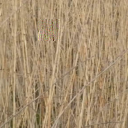}}
		\subfigure{\includegraphics[width=0.09\linewidth]{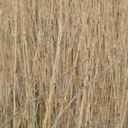}}
		\subfigure{\includegraphics[width=0.09\linewidth]{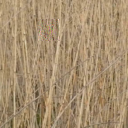}}
		\subfigure{\includegraphics[width=0.09\linewidth]{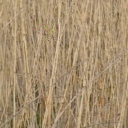}}\\
		\end{center}
	\vspace{-4mm}
		\caption{Dynamic texture inpainting using our method. We present the corrupted dynamic textures (1-st and 3-rd rows) and inpainted dynamic textures (2-nd and 4-th rows).}
		\label{dynamic_inpainting}
	\vspace{-4mm}
\end{figure*}

\begin{figure}[htb!]
		\begin{center}
		\subfigure[Corrupted]{
				\includegraphics[width=0.6\linewidth]{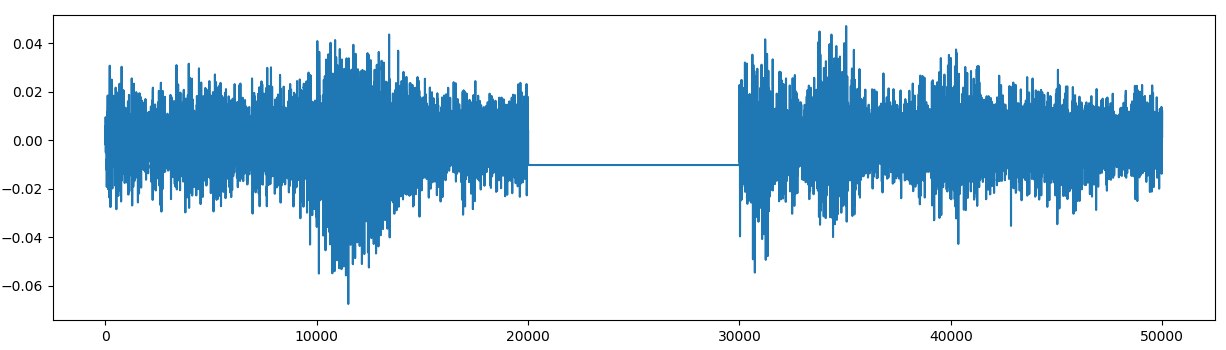}
		}\\
		\subfigure[Inpainted]{
				\includegraphics[width=0.6\linewidth]{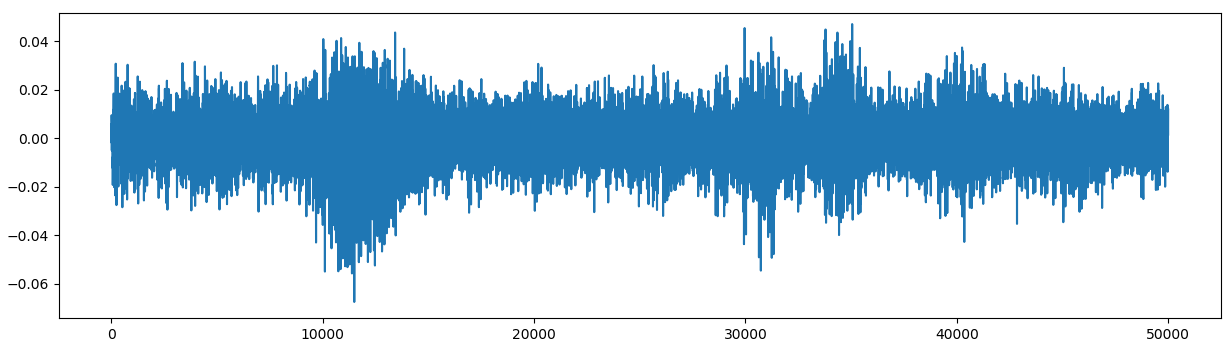}
		}
		\end{center}
	\vspace{-4mm}
		\caption{Sound texture inpainting using our method. Sound texture used here is ``bees''.}
		\label{Sound_inpainting}
	\vspace{-4mm}
\end{figure}

For image texture inpainting,
we evaluate our algorithm by comparing it with the following two deep image inpainting methods:
\begin{itemize}
	\item[-] {\textbf{Deep prior}}~\cite{ulyanov2018deep}: An inpainting algorithm that utilizes the prior of a random ConvNet.
	This method dose not require extra training data.
	\item[-] {\textbf{Deep fill}}~\cite{yu2018generative}: The state-of-the-art image inpainting algorithm. 
	It requires extra training data, and we use the model pretrained on ImageNet for our experiment.
\end{itemize}
We use the network $(4 \mathbf D \oplus 0 \mathbf S)$ where the number of channels is 64.
We first prepare a rectangle mask of size $(60, 60)$ near the center of a image,
then we obtain the corrupted texture by applying the mask to a raw texture,
\ie all pixels within the masked area are set to zero.
The border width is set to 4 pixels.
All inpainting methods have access to the mask and the corrupted texture,
but do not have access to the raw textures.

Fig.~\ref{image_inpainting} presents the qualitative comparison.
In general,
although the baseline methods can handle textures with non-local structures relatively well (the 1-st texture),
they can not handle random elements in textures (from the 2-nd to the 5-th textures).
Most results of baseline methods are blurry,
and the results of deep fill sometimes show obvious color artifacts (the 1-st and 5-th textures).
Clearly,
our method outperforms other baseline methods,
as it is able to inpaint all corrupted exemplars with convincing textural content,
and does not produce blurry or color artifacts.

Tab.~\ref{image_inpainting_quantitative} presents the quantitative comparison.
We calculate the MS-SSIM score between the inpainted textures and the corresponding raw textures (not shown).
A higher score indicates a better inpainting result.
It can be seen that our method outperforms other baseline methods in most cases.

For dynamic texture inpainting,
we prepare a mask of size $(25, 25)$,
and apply this mask to each frame of dynamic textures.
The border width is set to 2 pixels.
We use the network $(4 \mathbf D \oplus 0 \mathbf S)$ where the number of channels is reduced to 32.
The template is assigned by the user because the grid search may cause memory overhead for GPUs.
For sound texture inpainting,
the mask covers the interval from 20000-th to 30000-th data point.
The border width is set to 1000 data points.
We use the same network settings as in the sound texture synthesis experiment.

Fig.~\ref{dynamic_inpainting} and Fig.~\ref{Sound_inpainting} present the results of dynamic texture and sound texture inpainting using our method.
Similar to the case of image inpainting,
we observe that our method successfully fills the corrupted region with convincing textural content,
and the overall inpainted textures are natural and clear.
It should be noticed that our proposed method is the first neural algorithm for sound texture inpainting.

\section{Conclusion}
\label{conclusion}
In this paper, 
we present cgCNN for exemplar-based texture synthesis. 
Our model can synthesize high quality image texture, dynamic texture and sound textures in a unified manner. 
The experiments demonstrate the effectiveness of our model in texture synthesis, expansion and inpainting.

There are several issues need further investigations.
We notice that one limitation of cgCNN is that it cannot synthesis dynamic patterns without spatial stationarity,
such as the ones studied in~\cite{xie2017synthesizing}.
Extending cgCNN to those dynamic patterns would be an interesting direction for further work.
Another limitation is that current cgCNN can not learn multiple input textures,
\ie, it can only learn one texture a time.
Future works should extend cgCNN to the batch training setting, 
and explore its potential in down-stream tasks such as texture feature extraction~\cite{xia2017texture} and classification~\cite{xie2018learning}.

\bibliographystyle{IEEEtran}
\bibliography{egbib}
\end{document}